\documentclass[10pt,twocolumn,letterpaper]{article}
\usepackage{amsmath}

\usepackage{subcaption}
\usepackage{colortbl}  
\usepackage{booktabs}  
\usepackage{multirow}
\usepackage{graphicx}
\usepackage{subcaption}
\usepackage{adjustbox}
\usepackage{flushend}
\usepackage{csquotes}
\usepackage{geometry}
\usepackage[accsupp]{axessibility}  

\usepackage[pagenumbers]{cvpr} 

\definecolor{cvprblue}{rgb}{0.21,0.49,0.74}
\usepackage[pagebackref,breaklinks,colorlinks,allcolors=cvprblue]{hyperref}
\def\paperID{10363} 
\def\confName{CVPR}
\def\confYear{2025}

\title{COSMIC: Clique-Oriented Semantic Multi-space Integration for \\ Robust CLIP Test-Time Adaptation}

\author{
Fanding Huang\textsuperscript{1,\thanks{These authors contributed equally to this work.}}, 
Jingyan Jiang\textsuperscript{2,\footnotemark[1]}, 
Qinting Jiang\textsuperscript{1}, 
Hebei Li\textsuperscript{3}, 
Faisal Nadeem Khan\textsuperscript{1,\thanks{Co-corresponding authors.}}, 
\vspace{0.4cm}
Zhi Wang\textsuperscript{1,\footnotemark[2]}\\
\textsuperscript{1}Shenzhen International Graduate School, Tsinghua University\\
\textsuperscript{2}Shenzhen Technology University \;
\textsuperscript{3}University of Science and Technology of China
}

\begin{document}

\maketitle

\begin{abstract}
Recent vision-language models (VLMs) face significant challenges in test-time adaptation to novel domains. While cache-based methods show promise by leveraging historical information, they struggle with both caching unreliable feature-label pairs and indiscriminately using single-class information during querying, significantly compromising adaptation accuracy.
To address these limitations, we propose \textbf{COSMIC} (\underline{C}lique-\underline{O}riented \underline{S}emantic \underline{M}ulti-space \underline{I}ntegration for \underline{C}LIP), a robust test-time adaptation framework that enhances adaptability through multi-granular, cross-modal semantic caching and graph-based querying mechanisms. Our framework introduces two key innovations: \textit{Dual Semantics Graph} (DSG) and \textit{Clique Guided Hyper-class} (CGH). The Dual Semantics Graph constructs complementary semantic spaces by incorporating textual features, coarse-grained CLIP features, and fine-grained DINOv2 features to capture rich semantic relationships. Building upon these dual graphs, the Clique Guided Hyper-class component leverages structured class relationships to enhance prediction robustness through correlated class selection. Extensive experiments demonstrate COSMIC's superior performance across multiple benchmarks, achieving significant improvements over state-of-the-art methods: 15.81\% gain on out-of-distribution tasks and 5.33\% on cross-domain generation with CLIP RN-50. Code is available at \href{https://github.com/hf618/COSMIC}{github.com/hf618/COSMIC}.
\end{abstract}

\section{Introduction}
\label{sec:intro}

\begin{figure}[t]
    \centering
    \subfloat[Cache-based methods]{
        \includegraphics[width=0.8\linewidth]{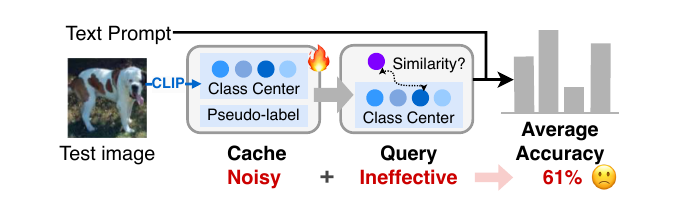}
    }
    \vfill
    \vspace{2mm}
    \subfloat[\centering Clique-{O}riented {S}emantic {M}ulti-space {I}ntegration for {C}LIP (Ours)]{
        \includegraphics[width=0.8\linewidth]{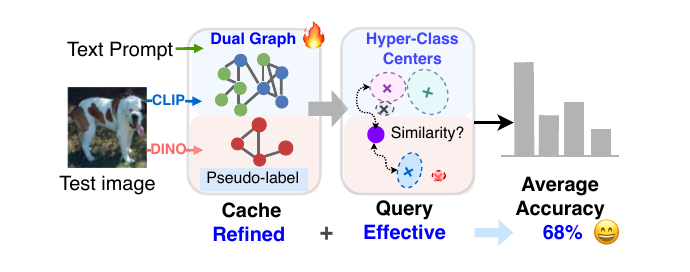}
    }
    \caption{(a) In the conventional cache-based method, the cache has only dull information with coarse-grained clip visual features and simple query way via similarity between samples and cached visual class centers. (b) In our COSMIC, the cache has diverse structural information via extra fine-grained DINOv2 visual features and effective query way via similarity between samples and meticulously designed hyper-class centers.}
    \label{motivation}
      \vspace{-0.3cm}
\end{figure}

Vision-language models (VLMs), such as CLIP~\cite{radford2021learning} and ALIGN~\cite{jia2021scaling}, have demonstrated remarkable performance across various downstream tasks, including semantic segmentation~\cite{li2022language, xu2022groupvit} and video understanding~\cite{tang2021clip4caption, xu2021videoclip}. This success can be attributed primarily to the alignment of visual and textual features on large-scale datasets, enabling these models to exhibit robust image understanding capabilities in open-world scenarios. However, when deployed in real-world applications, these models often encounter test samples that significantly deviate from the training dataset, resulting in performance degradation~\cite{zhou2022learning, zhou2022conditional}.

Recently, researchers~\cite{shu2022test} have explored the test-time adaptation (TTA) scenario for CLIP inference. TTA focuses on adapting models solely during testing by fine-tuning a small subset of the model's parameters, such as prompts (i.e., prompt learning), or even employing training-free cache-based methods to enhance the zero-shot image classification of CLIP. Prompt learning methods~\cite{shu2022test, feng2023diverse, abdul2024align} optimize textual learnable prompts by minimizing entropy with confidence selection, ensuring consistent predictions across various augmented views of each test image. Using the updated prompt, it generates adapted textual class features, making predictions based on their similarity to sample features. While effective, these approaches suffer from computational inefficiency due to numerous visual augmentations and iterative backpropagation steps. 

In contrast, as shown in Fig.~\ref{motivation} (a), cache-based methods~\cite{karmanov2024efficient,zhang2024dual} enhance model performance by utilizing historical information. Specifically, they propose an extra cache to store visual features with pseudo-labels to generate visual \textit{class centers} which are the average of previous visual features of each class. When querying the cache, features of the new test image are compared to find similar class centers. Then, labels corresponding to those centers are chosen to provide more information for the final prediction. Due to its training-free design and ability to leverage global historical information, the cache-based approach surpasses prompt learning in both effectiveness and efficiency.

However, two issues remain unaddressed in previous research: (1) Noisy pseudo-labels during cache construction contaminate the cache with unreliable feature-label pairs; (2) During querying, each class center encapsulates only single-class information, leading to blind propagation of a single pseudo-label regardless of its reliability. These dual flaws in cache construction and query mechanisms critically undermine adaptation accuracy.

To address the aforementioned issues, we introduce \textbf{COSMIC} (\underline{C}lique-\underline{O}riented \underline{S}emantic \underline{M}ulti-space \underline{I}ntegration for \underline{C}LIP), a robust CLIP test-time adaptation framework. Our key idea is to leverage multi-granularity and cross-modal semantic information to enrich the semantic content in the cache with limited samples while utilizing graph structures to organize and query the cache robustly. Specifically, as shown in Fig.~\ref{motivation} (b), we design two core components: (1) \textit{Dual Semantics Graph (DSG)} can enhance the semantic diversity of pseudo-labels by incorporating fine-grained DINOv2 visual features and cross-modal text features. Additionally, it bridges the gap between textual semantics and more granular visual semantics, thereby reducing the proportion of noisy feature and pseudo-label pairs in the cache.
(2) \textit{Clique Guided Hyper-class (CGH)} connects class centers and merges pseudo-labels to include information from multiple categories, thus improving the probability of containing correct label information in retrieved pseudo-labels. This, in turn, allows for selective use of this information in subsequent prediction phases, improving the model's robustness and accuracy.

\begin{figure}[t]
    \centering
    \subfloat[Different Visual Affinity.]{
        \includegraphics[width=0.47\linewidth]{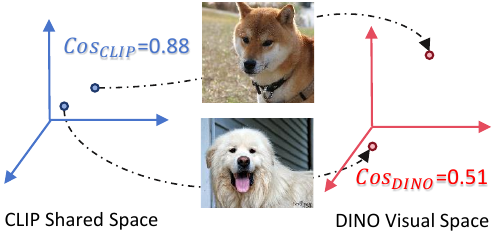}
    }
    \subfloat[Different Feature Cluster.]{
        \includegraphics[width=0.47\linewidth]{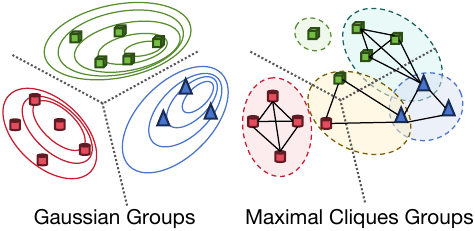}
    }
    \caption{(a) For various types of dogs, like \enquote{shiba inu} and \enquote{great Pyrenees} DINOv2 features offer more refined similarity perception than CLIP. (b) Samples in a single Gaussian cluster come from the same class, while the maximal clique cluster can explicitly represent cross-class correlations.}
    \label{fig:motivation2}
    \vspace{-0.6cm}
\end{figure}

For the Dual Semantics Graph, we initially construct complementary feature spaces with varying semantic granularity: (1) The \textit{CLIP Shared Semantic space} unifies text embeddings and visual class centers calculated by historical test features. (2) The \textit{Auxiliary Fine-grained Visual space} incorporates fine-grained visual class centers from self-supervised DINOv2~\cite{oquab2023dinov2}. As shown in Fig.~\ref{fig:motivation2} (a), CLIP's visual discrimination is less refined than DINOv2's for subtly varied pet images.   To capture intricate feature dynamics, we transform dual Euclidean spaces into dual graph spaces, modeling nonlinear interactions between classes.

For the Clique Guided Hyper-class, we design hyper-classes within each graph to represent latent semantics from various classes. Unlike previous approaches using Gaussian distributions~\cite{han2024dota}, as shown in Fig.~\ref{fig:motivation2} (b), our graph-based feature modeling facilitates robust relationships with fewer samples and effectively clusters samples from diverse classes. Therefore, we can efficiently search for maximal cliques to form the Clique Guided Hyper-class, which represents the centroids of diverse, highly-affiliated class centers in the dual graph. Then test features query the hyper-class centers by similarity to select highly correlated classes, referred to as inlier classes. Based on the logits of these inlier classes, we generate adapted predictions for test-time adaptation, ensuring a more robust and accurate model performance.

In summary, our contributions are as follows:

\begin{itemize}
    \item We introduce \textit{COSMIC}, a training-free test-time adaptation method for CLIP that explicitly captures complementary semantic relationships.
    \item To refine cache construction, we design a \textit{Dual Semantics Graph} that integrates intra-modal, cross-modal, coarse-grained, and fine-grained semantics.
    \item To query cache adaptively, we design \textit{Clique Guided Hyper-Class} to represent class clusters with high affinity, enabling more robust querying of test samples.

\end{itemize}

\vspace{-0.30cm}

\section{Related Work}
\label{sec:work}

\begin{figure*}[htbp]
    \centering
    \includegraphics[width=\textwidth]{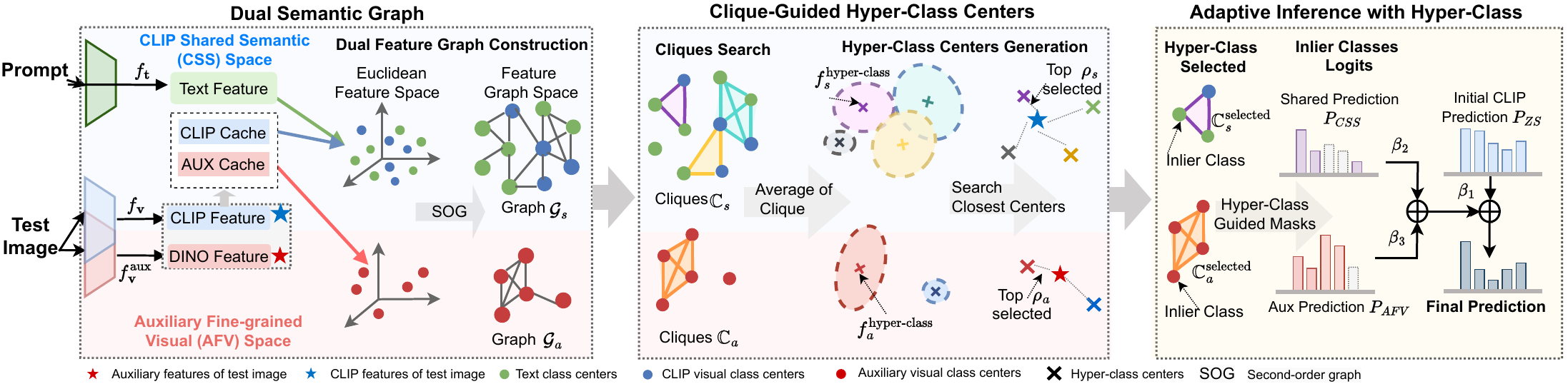}
    \caption{\textbf{Overview of COSMIC.} To refine cache with cross-modal, multi-granular class features, we construct \textit{Dual Semantics Graph} with complementary semantics, incorporating both joint modalities and fine-grained visual information. To efficiently query the compatibility of diverse semantics, we propose novel \textit{Clique Guided Hyper-class} to model different communities in the cache as the test domain evolves, enabling adaptive querying of test samples.}
    \label{fig:pdfimage}
  \vspace{-0.3cm}
\end{figure*}

\subsection{Vision-Language Models}

Pre-trained on large and high-quality datasets, vision-language models like CLIP~\cite{radford2021learning} achieve strong generalization by aligning textual and visual features via contrastive learning. For few-shot image classification, prompt learning improves adaptability by optimizing learnable tokens on the textual~\cite{zhou2022learning,zhou2022conditional} or viusal~\cite{khattak2023maple} encoders.

To avoid backpropagation costs, methods like Tip-Adapter~\cite{zhang2021tip}, CaFo~\cite{zhang2023CaFo} and GraphAdapter~\cite{li2024graphadapter} use vision adapters for VLM generalization. Although these methods enrich feature semantics in few-shot settings, we focus on zero-shot test-time adaptation, rigorously exploring cross-modal and multi-granular class centers to address the nonlinear relationships in new domains, better aligning with real-world deployment scenarios.

\vspace{-0.1cm}
\subsection{Test Time Adaptation}

Test-time adaptation addresses distribution shifts between training and test data. Recent approaches for VLMs include: TPT~\cite{shu2022test} leveraging prompt tuning for view consistency; SwapPrompt~\cite{ma2024swapprompt} employing self-contrastive learning for rapid adaptation; and PromptAlign~\cite{abdul2024align} aligning cross-domain statistical features. 

In contrast, cache-based methods like TDA~\cite{karmanov2024efficient} and DMN~\cite{zhang2024dual} construct dynamic caches to retrieve adapted visual class centers during testing. While this method enables CLIP to generalize via historical samples, such caches rely solely on CLIP's coarse text-image aligned features—lacking vision-specific contrastive knowledge critical for similarity awareness, as demonstrated by DINOv2~\cite{oquab2023dinov2}. Our work bridges this gap by enhancing cached features with textual and fine-grained visual semantics.

\vspace{-0.1cm}
\subsection{Pre-training of Visual Encoders} 
Modern vision systems predominantly rely on a two-stage paradigm: large-scale pretraining followed by task-specific fine-tuning. Despite its effectiveness, this approach poses limitations due to label dependency. Self-supervised learning mitigates this bottleneck by learning annotation-free representations, achieving state-of-the-art performance in semantic segmentation~\cite{huang2023dtbs, hoyer2022daformer} and depth estimation~\cite{hui2022rm}.

Transitioning from multimodal to view-consistent learning overcomes critical visual granularity limitations. While CLIP~\cite{radford2021learning} excels in cross-modal alignment through vision-language pretraining, its text-centric objective compromises spatial details vital for dense prediction. DINOv2~\cite{oquab2023dinov2} resolves this by learning augmentation-invariant features through self-distillation, preserving pixel-level nuances for annotation-free localization without textual guidance.
\section{Method}
\label{sec:method}

\subsection{Preliminaries}

\textbf{CLIP} model consists of two parallel encoders: a text encoder $F_t$ and a visual encoder $F_v$, which map features from different modalities into a shared embedding space. In a zero-shot $K$ class classification task, given a test image $\boldsymbol{x_\text{test}}$, the visual feature $\boldsymbol{w_\mathbf{v}}=F_v(\boldsymbol{x_\text{test}})$ serves as a query to the text features of all class descriptions. The probability that $\boldsymbol{x_\text{test}}$ belongs to the class $k$ can then be expressed as

\begin{equation}
    p^\mathrm{zs}_k(\boldsymbol{x_\text{test}})=\frac{\exp(\cos(\boldsymbol{w_\mathbf{v}},f_{\mathbf{t}_k})/\tau)}{\sum_{i=1}^K\exp(\cos(\boldsymbol{w_\mathbf{v}},f_{\mathbf{t}_i})/\tau)},
\end{equation}
where $f_{\mathbf{t}_i}$ represents the textual class feature for the $i^{th}$ class, and $\tau$ denotes the temperature parameter in the softmax function.

Inspired by few-shot adaptation of CLIP method Tip-Adapter~\cite{zhang2021tip}, \textbf{TDA}~\cite{karmanov2024efficient} is a typical cache-based method for TTA that utilizes a cache to store knowledge from high-confidence historical test samples. This cache operates as a queue-style dictionary, where the keys represent the pseudo-labels of the samples, and the values store the corresponding visual features. The decision to add a sample to the cache depends on its prediction entropy. Specifically, for a test sample $\boldsymbol{x_\text{test}}$, after undergoing random $\mathcal{N}$ augmentations $\mathcal{A}$ (including the original image), the marginal entropy of the sample predictions is calculated as:
\vspace{-2.5mm}
\begin{equation}
\begin{aligned}
    H(\boldsymbol{x}_\text{test}) &= -\sum_{i=1}^K \tilde{p}_i(\boldsymbol{x}_\text{test}) \log \tilde{p}_i(\boldsymbol{x}_\text{test}), \\
    \tilde{p}_i(\boldsymbol{x}_\text{test}) &= \frac{1}{\mathcal{N}_{\text{max}}}\sum_{j=1}^{\mathcal{N}_{\text{max}}} p_i^\mathrm{zs}(\mathcal{A}_j(\boldsymbol{x}_\text{test})),
\end{aligned}
\end{equation}
where $p_i^\mathrm{zs}(\mathcal{A}_j(\boldsymbol{x}_\text{test}))$ represents probability of class $i$ generated by the model for the $j^{th}$ augmented view of the test image. $N_{max}$ is the number of first $R \times \mathcal{N}$ high-confidence view predictions, where $R$ is the selection ratio. Based on this, the logic for storing $\boldsymbol{w_\mathbf{v}}$ predicted as class $i$ into cache set $\mathbb{M}_i$ is as follows: 
\begin{align}
\mathbb{M}_i = 
\begin{cases}
    \mathbb{M}_i \cup \boldsymbol{w_\mathbf{v}}, & \quad \text{if } |\mathbb{M}_i| < L, \\
    \mathbb{M}_i \setminus \{{f}_{\max}\} \cup \boldsymbol{w_\mathbf{v}}, & \enspace
    \begin{array}{l}
        \text{if } |\mathbb{M}_i| = L \text{ and } \\
        H(\boldsymbol{x_\text{test}}) < H_{\max}(\mathbb{M}_i),
    \end{array} \\
    \mathbb{M}_i, & \quad \text{otherwise}.
\end{cases}
\label{cache}
\end{align}

Here, ${f}_{\text{max}}$ denotes the feature in cached features set $\mathbb{M}_i$ with the highest entropy $H_{\max}(\mathbb{M}_i)$, and $L$ denotes the maximum capacity of the cache. Using this cache, the adapted predictions are computed as follows:
\begin{equation}
\mathbf{P}_\text{adapted}(\boldsymbol{x_\text{test}})=\varphi(\boldsymbol{w_\mathbf{v}}\mathbf{M}^\top)\mathbf{{L}_\text{pseudo}},
\label{adptation}
\end{equation}
where $\varphi(x)=\exp(-\alpha(1-x))$ is an adaptation function, $\mathbf{M} \in\mathbb{R}^{KL\times d} $ is the cached features matrix, and $\mathbf{{L}_\text{pseudo}} \in\mathbb{R}^{KL\times K}$ is the one-hot pseudo-label matrix.

\subsection{Dual Semantics Graph (DSG)}
\label{Anchors}

\subsubsection{CLIP Shared Semantic (CSS) Space}
\label{CLIP Shared Space}
CLIP aims to align text and visual features in a shared feature space. Leveraging this, we combine text features $\mathbf{f_t} = [f_{\mathbf{t}_1} f_{\mathbf{t}_2} \cdots f_{\mathbf{t}_K}]^\top \in \mathbb{R}^{K\times d_1}$ with cached visual class centers $\mathbf{f_v} = [f_{\mathbf{v}_1} f_{\mathbf{v}_2} \cdots f_{\mathbf{v}_K}]^\top \in \mathbb{R}^{K\times d_1}$ in the same Euclidean space, forming a unified feature set $\mathbf{f_s} =\mathbf{f_t}\cup\mathbf{f_v}= [f_{\mathbf{s}_1} f_{\mathbf{s}_2} \cdots f_{\mathbf{s}_{2K}}]^\top \in \mathbb{R}^{2K\times d_1}$, termed the CLIP Shared Semantic space with $d_1$ dimension. The text feature $f_{\mathbf{t}_i}$ are CLIP embeddings of class $i$ description, defined as:
\begin{equation}
    f_{\mathbf{t}_i}=F_t(\text{Text}_i),
\end{equation}
where $\text{Text}_i$ is the input text of class $i$. Inspired by DMN~\cite{zhang2024dual}, we construct the visual center $f_{\mathbf{v}_i}$ for class $i$ using a weighted combination of visual features, determined by the cosine similarity between test and cached visual features, illustrated as:
\vspace{-1mm}
\begin{equation}
    f_{\mathbf{v}_i} =\varphi\left(\boldsymbol{w_\mathbf{v}}\mathbf{M}_i^\top\right)\mathbf{M}_i,
\end{equation}
where $\mathbf{M}_i \in \mathbb{R}^{l_1 \times d_1}$ represents the cached visual feature matrix from the $i^{th}$ category with a capacity $l_1$ of cache.

\subsubsection{Auxiliary Fine-grained Visual (AFV) Space}
While the CLIP Shared Semantic space leverages both text and visual features for alignment, the cache model eliminates the need for text features during inference. However, relying solely on the CLIP visual encoder under noisy pseudo-labels can lead to inaccurate similarity perception, especially for easily confused images. Therefore, we introduce an auxiliary visual feature branch using self-supervised encoders like DINOv2~\cite{oquab2023dinov2} to generate more finer-grained visual features of $\boldsymbol{x_\text{test}}$.

Analogously to the CLIP visual feature cache, we establish a DINOv2 visual feature cache using the same criteria as in Eq.~\ref{cache}. To average historical test information with fine-grained semantics, we compute the auxiliary visual center $f_{\mathbf{v}_i}^\text{aux}$ via the centroid of cached DINOv2 features for class $i$:
\vspace{-1mm}
\begin{equation}
\begin{aligned}
    \mathbf{f_v^\text{aux}} &= \begin{bmatrix} f_{\mathbf{v}_1}^\text{aux} & f_{\mathbf{v}_2}^\text{aux} & \dots & f_{\mathbf{v}_C}^\text{aux} \end{bmatrix}^\top \in \mathbb{R}^{K\times d_2}, \\
    f_{\mathbf{v}_i}^\text{aux}&=\frac{1}{l_2} \sum_{j=1}^{l_2}\mathbf{M}^\text{aux}_i[j,:],
\end{aligned}
\end{equation}
where $l_2$ and $d_2$ are the capacity of the auxiliary cache and dimension of AFV space, respectively. $\mathbf{M}^\text{aux}_i \in \mathbb{R}^{l_2 \times d_2}$ is the auxiliary cached visual feature matrix for the $i^{th}$ class. 

\subsubsection{Second-Order Graph Construction}
In contrast to Euclidean space, graph space provides a more precise representation of complex affinity relationships~\cite{yao2025mecon, zhang20233d} between intra- and inter-modal features. Specifically, we construct two types of graphs: the First-Order Graph (FOG), which captures direct pairwise similarities, and the Second-Order Graph (SOG), which extends this by modeling higher-order relationships through the product of adjacency matrices, enabling the discovery of more complex and hidden patterns. In each FOG, nodes represent class center features, and edges capture bi-directional compatibility. Let $\mathbf{F}\in\mathbb{R}^{N\times D}$ represent the feature matrix, where $N=2K,D=d_1$ for CSS space and $N=K,D=d_2$ for AFV space. The adjacency matrix of First-Order Graph is defined as:
\vspace{-1.5mm}
\begin{equation}
    \mathcal{W}_{FOG}=[w_{ij}]_{N\times N},\; w_{ij}=
    \begin{cases}
        1 & \text{if }\frac{\mathbf{F}_i\cdot\mathbf{F}_j}{\|\mathbf{F}_i\|\|\mathbf{F}_j\|}>t^\text{aff} \\
        0 & \mathrm{otherwise}
    \end{cases},
\end{equation}
where $t^\text{aff}$ is a threshold changed by test iterations. Overly concentrated class centers in specific domains can lead to excessive redundancy in graph information. To address this, the Second-Order Graph (SOG) captures higher-order dependencies through the adjacency matrix:
\begin{equation}
    \mathcal{W}_{SOG}=\mathcal{W}_{FOG}\odot(\mathcal{W}_{FOG}\times\mathcal{W}_{FOG}). 
\end{equation}

Unlike FOG, SOG models indirect relationships, such as transitive or multi-hop connections, providing a more comprehensive view of node interactions. Additionally, the SOG is sparser than the FOG, reducing noise and improving computational efficiency. As testing progresses, the CSS and AFV graphs accumulate higher-quality features from new domains. To adapt to this dynamic, we adopt an increasing rule for the threshold $t^\text{aff}_i=\min\left(1,t^\text{aff}_0+g\cdot i\right)$ where $t^\text{aff}_0$ is the initial value, $i$ is the current number of samples tested, and $g$ is a constant growth rate. 

\subsection{Clique Guided Hyper-class (CGH)}

\paragraph{Maximal Cliques Search.} 
Given an undirected graph $\mathcal{G} = (\mathcal{V},\mathcal{E})$, clique $\mathcal{C}=\mathcal{(V^{\prime},E^{\prime}),V^{\prime}\subseteq V,E^{\prime}\subseteq E}$ is a complete subgraph of $\mathcal{G}$. A maximal clique is a clique that cannot be extended by including any additional node. We employ the modified Bron-Kerbosch algorithm~\cite{eppstein2010listing} for maximal clique search, which is a highly efficient algorithm with a worst-case time complexity of $\mathcal{O}(b(n - b)3^{(b/3)})$, where $b$ denotes the graph’s degeneracy. When the $t^\text{aff}$ is high, the graph becomes sparser, leading to a lower degeneracy, which further accelerates the search process. For CSS and AFV spaces, we define their graphs:
\begin{equation}
    \mathcal{G}_\text{CSS} = (\mathcal{V}_\text{CSS}, \mathcal{E}_\text{CSS}) \text{, } \mathcal{G}_\text{AFV} = (\mathcal{V}_\text{AFV}, \mathcal{E}_\text{AFV}),
\end{equation}
where $\mathcal{G}_\text{CSS}$ and $\mathcal{G}_\text{AFV}$ represent the graphs in the CSS and AFV spaces, respectively.  For each graph $j \in \{\text{CSS}, \text{AFV}\}$, we apply the modified Bron-Kerbosch algorithm to both graphs:
\vspace{-4mm}

\begin{equation}
    \mathbb{C}_j = \text{BronKerbosch}(\mathcal{G}_j),
\end{equation}
where $\mathbb{C}_j=\{\mathcal{C}_{j1}, \mathcal{C}_{j2}, ..., \mathcal{C}_{jm_j}\}$ is the set of maximal cliques in graph $\mathcal{G}_j$. These two sets of maximal cliques, $\mathbb{C}_\text{CSS}$ and $\mathbb{C}_\text{AFV}$, enable us to capture a more comprehensive understanding of feature affinities across diverse levels of semantics and granularities.
\vspace{-4mm}

\paragraph{Hyper-classes Generation.}
\label{Dual-Space Cliques Guided Anchors}
Building on the maximal cliques identified in both $\mathcal{G}_\text{CSS}$ and $\mathcal{G}_\text{AFV}$, we further reveal latent patterns by searching for \textit{Clique-Guided Hyper-classes}. Given the dense connectivity within maximal cliques, we define hyper-class centers as the centroids of node classes within clique $i$.
\vspace{-0.5mm}
\begin{equation}
   f^{\text{hyper-class}}_{ji} =\frac1{|\mathcal{C}_{ji}|}\sum_{f\in \mathcal{C}_{ji}}f,
\end{equation}
where $f$ represents class nodes in the respective graph. This approach models community structures between class centers through graph-based contextual reasoning, overcoming the locality bias of similarity-query paradigms. The affinity between test image feature $\boldsymbol{w_v}$ and clique $i$ in $\mathcal{G}_j$ is:

\vspace{-2mm}
\begin{equation}
    \rho_j(\boldsymbol{w_\mathbf{v}},\mathcal{C}_{ji})=\text{cos}\left(\boldsymbol{w_\mathbf{v}},f^{\text{hyper-class}}_{ji}\right).
\end{equation}

To identify the hyper-classes most proximal to the test sample, we sort the cliques by affinity: $\rho_j(\boldsymbol{w_\mathbf{v}},\mathcal{C}_{j(1)}) \geq \rho_j(\boldsymbol{w_\mathbf{v}},\mathcal{C}_{j(2)}) \geq \cdots \geq \rho_j(\boldsymbol{w_\mathbf{v}},\mathcal{C}_{j(m_j)})$. Then we select the top $r$ proportion of the closest hyper-classes in each graph:

\begin{equation}
    \mathbb{C}_{j}^{\text{selected}}=\{\mathcal{C}_{j(1)},\mathcal{C}_{j(2)},\ldots,\mathcal{C}_{j(k_j)}\},
\end{equation}
where $k_j=\lceil r\cdot m_j\rceil$. This procedure generates two hyper-class guided masks $\mathcal{M}_{j}$, effectively delineating the selected classes as inliers while designating the remainder as outliers (classes with low relevance to the test sample).

\vspace{-6mm}

\begin{equation}
\begin{aligned}
    \mathcal{M}_{j} &= \begin{bmatrix} m_{j1} & m_{j2} & \dots & m_{jN_j} \end{bmatrix} \in \mathbb{R}^{1 \times N_j}, \\
    \quad m_{ji} &= 
    \begin{cases} 
    1, & \text{if } i \in \bigcup_{\mathcal{C} \in \mathbb{C}_{j}^{\text{selected}}} \mathcal{C} \\ 
    0, & \text{otherwise} 
    \end{cases},
\end{aligned}
\end{equation}
where $N_\text{CSS} = 2K$ for the CLIP Shared Semantic space and $N_\text{AFV} = K$ for the Auxiliary Fine-grained Visual space.

\subsection{Adaptive Inference with Hyper-class}
The $\mathcal{M}_\text{CSS}$ is then applied to the initial logits of all classes in the CSS space:
\begin{equation}
\mathbf{P}_\text{CSS}^{\text{initial}}=\text{Softmax}(\boldsymbol{w_\mathbf{v}}\mathbf{f_s}^\top) \odot \mathcal{M}_\text{CSS}\in \mathbb{R}^{1\times 2K}.
\end{equation}

The probability $p_i$ for the $i^{th}$ class in the CSS space is calculated as the average of the text and image class predictions:
\vspace{-2mm}
\begin{equation}
    p_i=(p_{i}^{\text{initial}}+p_{i+K}^{\text{initial}})/2, \quad\mathrm{for}\quad i=1,2,\ldots,K.
\end{equation}

Let $p_i^{\text{initial}}$ denote the prediction value in the $i^{th}$ column of $\mathbf{P}_\text{CSS}^{\text{initial}}$. The classification probability distribution in the CSS space is then given by:
\begin{equation}
    \mathbf{P}_\text{CSS}=\begin{bmatrix} p_1 & p_2 & \dots & p_K \end{bmatrix} \in \mathbb{R}^{1\times K},
\end{equation}
where $K$ is the number of classes. Following a similar process, we obtain the class prediction probability distribution within the AFV space:
\begin{equation}
    \mathbf{P}_\text{AFV}=\text{Softmax}(\boldsymbol{w_{\mathbf{v}}^\text{aux}}{\mathbf{f_v^\text{aux}}}^\top) \odot \mathcal{M}_\text{AFV} \in \mathbb{R}^{1\times K},
\end{equation}
where $\boldsymbol{w_{\mathbf{v}}^\text{aux}}$ is the auxiliary image feature of test image $\boldsymbol{x_\text{test}}$. The calculation of the $\mathcal{M}_\text{AFV}$ is similar to $\mathcal{M}_\text{CSS}$, but they mainly filter out outliers from various feature clusters, thereby preventing prediction bias caused by the accumulation of moderate logits values. We combine initial CLIP prediction with adaptive prediction into the final prediction:

\begin{equation}
\begin{aligned}
    \mathbf{P}_\text{Final}&=\beta_1\mathbf{P}_\text{ZS}+\beta_2\mathbf{P}_\text{CSS}+\beta_3\mathbf{P}_\text{AFV}, \\
     \mathbf{P}_\text{ZS}&=\text{Softmax}(\boldsymbol{w_\mathbf{v}}\mathbf{f_t})^\top\in \mathbb{R}^{1\times K},
\end{aligned}
\label{add eq}
\end{equation}
where $\beta_1,\beta_2,\beta_3$ represent the weights, and their values will be discussed in detail in Sec.~\ref{betas}.

\begin{table*}[htbp]
  \centering
  \caption{Top-1 accuracy (\%) comparison on ImageNet and its OOD variants using CLIP with ResNet-50 and ViT-B/16 backbones. Our results are reported as mean$\pm$std over 3 random seeds. \textbf{Bold} indicates the highest performance.}
    \resizebox{\linewidth}{!}{
    \begin{tabular}{llccccccc}
    \toprule
    Method & Adaptation Settings & ImageNet & ImageNet-A & ImageNet-V2 & ImageNet-R & ImageNet-S & Average & OOD Average \\
    \midrule
    CLIP-RN-50 & -     & 58.16  & 21.83  & 51.41  & 56.15  & 33.37  & 44.18  & 40.69  \\
    \midrule
    CoOp~\cite{zhou2022learning} (IJCV'22)  & Training Few-shot & 63.33  & 23.06  & 55.40  & 56.60  & 34.67  & 46.61  & 42.43  \\
    CoCoOp~\cite{zhou2022conditional} (CVPR'22) & Training Few-shot & 62.81  & 23.32  & 55.72  & 57.74  & 34.48  & 46.81  & 42.82  \\
    TPT~\cite{shu2022test} (NeurIPS'22)  & Training Zero-shot & 60.74  & 26.67  & 54.70  & 59.11  & 35.09  & 47.26  & 43.89  \\
    DiffTPT~\cite{feng2023diverse} (ICCV'23) & Training Zero-shot & 60.80  & 31.06  & 55.80  & 58.80  & 37.10  & 48.71  & 45.69  \\
    TDA~\cite{karmanov2024efficient} (CVPR'24) & Training-free Zero-shot & 61.35  & 30.29  & 55.54  & 62.58  & 38.12  & 49.58  & 46.63  \\
    DMN~\cite{zhang2024dual} (CVPR'24) & Training-free Zero-shot & 63.87  & 28.57  & 56.12  & 61.44  & 39.84  & 49.97  & 46.49  \\
\rowcolor[rgb]{ .949,  .949,  .949} COSMIC (Ours)  & Training-free Zero-shot & \textbf{75.19}±0.89 & \textbf{49.07}±0.06 & \textbf{63.81}±0.39 & \textbf{79.87}±0.08 & \textbf{56.44}±0.34 & \textbf{64.88}±0.22 & \textbf{62.30}±0.05 \\
    \bottomrule
    \toprule
    Method & Adaptation Settings & ImageNet & ImageNet-A & ImageNet-V2 & ImageNet-R & ImageNet-S & Average & OOD Average \\
    \midrule
    CLIP-ViT-B/16 & -     & 66.73  & 47.87  & 60.86  & 73.98  & 46.09  & 59.11  & 57.20  \\
    \midrule
    CoOp~\cite{zhou2022learning} (IJCV'22) & Training Few-shot & 71.51  & 49.71  & 64.20  & 75.21  & 47.99  & 61.72  & 59.28  \\
    CoCoOp~\cite{zhou2022conditional} (CVPR'22) & Training Few-shot & 71.02  & 50.63  & 64.07  & 76.18  & 48.75  & 62.13  & 59.91  \\
    TPT~\cite{shu2022test} (NeurIPS'22) & Training Zero-shot & 68.98  & 54.77  & 63.45  & 77.06  & 47.94  & 62.44  & 60.81  \\
    DiffTPT~\cite{feng2023diverse} (ICCV'23) & Training Zero-shot & 70.30  & 55.68  & 65.10  & 75.00  & 46.80  & 62.58  & 60.65  \\
    TDA~\cite{karmanov2024efficient} (CVPR'24) & Training-free Zero-shot & 69.51  & 60.11  & 64.67  & 80.24  & 50.54  & 65.01  & 63.89  \\
    DMN~\cite{zhang2024dual} (CVPR'24) & Training-free Zero-shot & 72.25  & 58.28  & 65.17  & 78.55  & 53.20  & 65.49  & 63.80  \\
\rowcolor[rgb]{ .949,  .949,  .949} COSMIC (Ours)  & Training-free Zero-shot & \textbf{78.19}±0.56 & \textbf{73.32}±0.32 & \textbf{69.62}±0.19 & \textbf{85.60}±0.12 & \textbf{62.79}±0.10 & \textbf{73.90}±0.05 & \textbf{72.83}±0.10 \\
    \bottomrule
    \end{tabular}%
      }
  \label{OOD_SOTA}
\end{table*}%
 
  \section{Experiments}
\label{sec:Experiments}

\subsection{Experimental Settings}

\paragraph{Datasets.} In the VLM test-time adaptation setting, two main benchmark types are typically used. The first evaluates the model's robustness under out-of-distribution (OOD) shifts, while the second examines cross-domain (CD) generalization capabilities.

\begin{itemize}
    \item For OOD shifts, we employ the ImageNet validation~\cite{deng2009imagenet} along with four variants: ImageNet-A~\cite{hendrycks2021natural}, ImageNet-V2~\cite{recht2019imagenet}, ImageNet-R~\cite{hendrycks2021many}, and ImageNet-Sketch~\cite{wang2019learning}.
    \item For CD tasks, we utilize ten diverse sub-datasets, each representing a distinct domain: Aircraft~\cite{maji2013fine}, Caltech101~\cite{fei2004learning}, Cars~\cite{krause20133d}, DTD~\cite{cimpoi2014describing}, EuroSAT~\cite{helber2019eurosat}, Flower102~\cite{nilsback2008automated}, Food101~\cite{bossard2014food}, Pets~\cite{parkhi2012cats}, SUN397~\cite{xiao2010sun}, and UCF101~\cite{khurram2012ucf101}.
\end{itemize}

\paragraph{Implement Settings.} We employed CLIP's officially pre-trained vision encoders, including ResNet50~\cite{he2016deep} and ViT-B/16~\cite{dosovitskiy2020image}. Inspired by previous works~\cite{zhang2024dual}, we utilized handcrafted textual prompts and multi-view augmentations of the original images to calculate prediction confidence. For all experiments, we generated 16 views per image and fixed hyperparameters across each sub-dataset. To combine logits, we used an adaptive step search to determine optimal weights while also demonstrating the robustness of fixed weights. Unless otherwise specified, we employ DINOv2 ViT-L/14 as the auxiliary visual encoder. All experiments were conducted on a single Tesla V100S-PCIE-32GB GPU, using top-1 accuracy for classification performance.

\vspace{-3mm}

\paragraph{Comparison Methods.} We initially evaluate few-shot methods such as CoOp~\cite{zhou2022learning} and CoCoOp~\cite{zhou2022conditional}, using 16-shot annotated samples per class. Next, we assess training-based zero-shot methods. TPT~\cite{shu2022test} straightforwardly optimizes the text prompts by minimizing the multi-view marginal entropy. DiffTPT~\cite{feng2023diverse} is an advanced iteration of TPT, employing diffusion-based augmentations to refine prompts. Furthermore, we evaluate training-free zero-shot methods. TDA~\cite{karmanov2024efficient} is an adapter-based approach that builds positive and negative caches at test time without training. DMN~\cite{zhang2024dual} leverages a dynamic memory to compile information from past test data, bypassing backpropagation.

\begin{table*}[htbp]
  \centering
  \caption{Top-1 accuracy (\%) comparison on 10 diverse cross-domain datasets using CLIP with ResNet-50 and ViT-B/16 backbones. Our results are reported as mean$\pm$std over 3 random seeds. \textbf{Bold} indicates the highest performance.}
    \resizebox{\textwidth}{!}{
    \begin{tabular}{llccccccccccc}
    \toprule
    Method & Adaptation Settings & Aircraft & Caltech101 & Cars  & DTD   & EuroSAT & Flower102 & Food101 & Pets  & SUN397 & UCF101 & Average \\
    \midrule
    CLIP-RN-50 & -     & 15.66  & 85.88  & 55.70  & 40.37  & 23.69  & 61.75  & 73.97  & 83.57  & 58.80  & 58.84  & 55.82  \\
    \midrule
    CoOp~\cite{zhou2022learning} (IJCV'22)& Training Few-shot & 15.12  & 86.53  & 55.32  & 37.29  & 26.20  & 61.55  & 75.59  & 87.00  & 58.15  & 59.05  & 56.18  \\
    CoCoOp~\cite{zhou2022conditional} (CVPR'22) & Training Few-shot & 14.61  & 87.38  & 56.22  & 38.53  & 28.73  & 65.57  & 76.20   & 88.39  & 59.61  & 57.10  & 57.23  \\
    TPT~\cite{shu2022test} (NeurIPS'22) & Training Zero-shot & 17.58  & 87.02  & 58.46  & 40.84  & 28.33  & 62.69  & 74.88  & 84.49  & 61.46  & 60.82  & 57.66  \\
    DiffTPT~\cite{feng2023diverse} (ICCV'23) & Training Zero-shot & 17.60  & 86.89  & 60.71  & 40.72  & 41.04  & 63.53  & 79.21  & 83.40  & 62.72  & 62.67  & 59.85  \\
    TDA~\cite{karmanov2024efficient} (CVPR'24)  & Training-free Zero-shot & 17.61  & 89.70  & 57.78  & 43.74  & 42.11  & 68.74  & 77.75  & 86.18  & 62.53  & 64.18  & 61.03  \\
    DMN~\cite{zhang2024dual} (CVPR'24) & Training-free Zero-shot & 22.77  & 90.14  & 60.02  & 50.41  & 48.72  & 67.93  & 76.70  & 86.78  & 64.39  & 65.34  & 63.32  \\
\rowcolor[rgb]{ .949,  .949,  .949} COSMIC (Ours)  & Training-free Zero-shot & \textbf{25.49}±0.57 & \textbf{94.77}±0.35 & \textbf{66.69}±0.51 & \textbf{55.44}±0.68 & \textbf{48.97}±2.41 & \textbf{77.78}±1.22 & \textbf{83.53}±0.11 & \textbf{92.63}±0.29 & \textbf{69.73}±0.08 & \textbf{71.51}±0.78 & \textbf{68.65}±0.32 \\
    \bottomrule
    \toprule
    Method & Adaptation Settings & Aircraft & Caltech101 & Cars  & DTD   & EuroSAT & Flower102 & Food101 & Pets  & SUN397 & UCF101 & Average \\
    \midrule
    CLIP-ViT-B/16 & -     & 23.22  & 93.55  & 66.11  & 45.04  & 50.42  & 66.99  & 82.86  & 86.92  & 65.63  & 65.16  & 64.59  \\
    \midrule
    CoOp~\cite{zhou2022learning} (IJCV'22)   & Training Few-shot & 18.47  & 93.70  & 64.51  & 41.92  & 46.39  & 68.71  & 85.30  & 89.14  & 64.15  & 66.55  & 63.88  \\
    CoCoOp~\cite{zhou2022conditional} (CVPR'22) & Training Few-shot & 22.29  & 93.79  & 64.90  & 45.45  & 39.23  & 70.85  & 83.97  & 90.46  & 66.89  & 68.44  & 64.63  \\
    TPT~\cite{shu2022test} (NeurIPS'22)  & Training Zero-shot & 24.78  & 94.16  & 66.87  & 47.75  & 42.44  & 68.98  & 84.67  & 87.79  & 65.50  & 68.04  & 65.10  \\
    DiffTPT~\cite{feng2023diverse} (ICCV'23) & Training Zero-shot & 25.60  & 92.49  & 67.01  & 47.00  & 43.13  & 70.10  & \textbf{87.23}  & 88.22  & 65.74  & 62.67  & 64.92  \\
    TDA~\cite{karmanov2024efficient} (CVPR'24) & Training-free Zero-shot & 23.91  & 94.24  & 67.28  & 47.40  & 58.00  & 71.42  & 86.14  & 88.63  & 67.62  & 70.66  & 67.53  \\
    DMN~\cite{zhang2024dual} (CVPR'24)   & Training-free Zero-shot & 30.03  & 95.38  & 67.96  & 55.85  & \textbf{59.43}  & 74.49  & 85.08  & 92.04  & 70.18  & 72.51  & 70.30  \\
    \rowcolor[rgb]{ .949,  .949,  .949} COSMIC (Ours)  & Training-free Zero-shot & \textbf{31.44}±0.56 & \textbf{96.80}±0.42 & \textbf{71.31}±0.46 & \textbf{58.23}±1.40 & 58.82±0.40 & \textbf{82.14}±0.49 & 86.60±0.09 & \textbf{94.19}±0.09 & \textbf{72.33}±0.06 & \textbf{76.20}±0.55 & \textbf{72.81}±0.09 \\
    \bottomrule

    \end{tabular}%
  }
  \label{CD_SOTA}%
\end{table*}%

\begin{figure*}[t] 
  \centering
  \includegraphics[width=\textwidth]{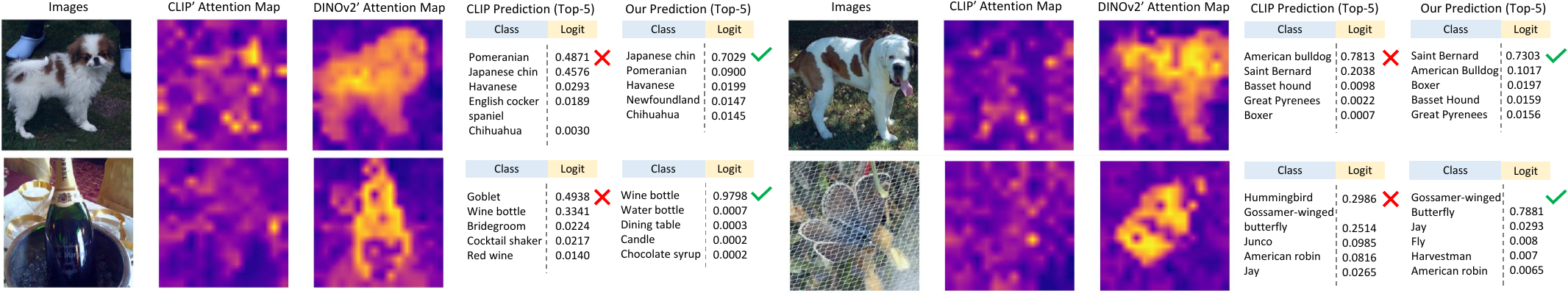} 
  \caption{Visualization of the attention maps from CLIP ViT-B/16 and DINOv2 ViT-B/14, along with their top-5 prediction results.}
  \label{fig:attn}
  \vspace{-0.3cm}
\end{figure*}

\vspace{-2mm}

\subsection{Comparison with SOTA}

\subsubsection{Results on the OOD Benchmark}

In Tab.~\ref{OOD_SOTA}, we compare model performance on the in-domain ImageNet validation set and four OOD variants. Due to domain shifts, the zero-shot generalization ability of CLIP is limited in OOD scenarios. Notably, while CoOp~\cite{zhou2022learning} and CoCoOp~\cite{zhou2022conditional} enhance CLIP's transferability by fine-tuning prompts on few-shot samples, this incurs additional training costs, making it impractical for real-world deployment. In contrast, our proposed method is training-free, allowing for immediate adaptation to unseen domains. Furthermore, our method achieves substantial improvements over CoCoOp, with gains of 19.48\% and 12.92\% across different CLIP visual encoders, demonstrating its superior performance in OOD generalization tasks.

Additionally, our method outperforms previous test-time adaptation techniques. For prompt-learning-based methods, our approach surpasses TPT~\cite{shu2022test} and DiffTPT~\cite{feng2023diverse}, with improvements of 18.41\%, 12.02\%, and 16.61\%, 12.18\% on two different backbones with OOD shifts, respectively, highlighting the effectiveness of our training-free approach. In comparison to cache model-based methods, our approach achieves superior results over the strongest techniques, TDA~\cite{karmanov2024efficient} and DMN~\cite{zhang2024dual}, with gains of 15.67\%, 15.81\%, and 8.94\%, 9.03\% on two backbones with OOD shifts, respectively, validating the effectiveness of our Dual Semantics Graph and Clique Guided Hyper-class to improve the refinement and querying in cache.

\begin{table}[htbp]
  \centering
  \caption{Efficiency and effectiveness comparison of TTA methods on Flower102~\cite{nilsback2008automated} dataset using CLIP ViT-B/16.}
  \resizebox{\columnwidth}{!}{%
    \begin{tabular}{lcccc}
    \toprule
    Method & Views & Time Per Image (s) & Top-1 Accuracy (\%) & Gain (\%) \\
    \midrule
    CLIP~\cite{radford2021learning} & 1 & 0.098 & 66.99 & - \\
    TPT~\cite{shu2022test}   & 64    & 0.198 & 68.98 & 1.99 \\
    TDA~\cite{karmanov2024efficient}   & 64    & \textbf{0.119} & 71.42 & 4.43 \\
    DMN~\cite{zhang2024dual}   & 128   & 0.483 & 74.49 & 7.50 \\
    \rowcolor[rgb]{ .949,  .949,  .949} COSMIC  & \textbf{16}    & 0.368 & \textbf{82.46} & \textbf{15.47} \\
    \bottomrule
    \end{tabular}%
      } 
  \label{eff}%
  \vspace{-0.3cm}
\end{table}

\subsubsection{Results on the CD Benchmark}

We conducted a comparison on a diverse cross-domain dataset against a variety of contemporary methods, as shown in Tab.~\ref{CD_SOTA}. Notably, both few-shot and zero-shot prompt learning methods show limited improvements in generalization performance. This limitation arises because, when there is a significant shift both in textual and visual modality, it becomes challenging to identify an optimal prompt within a constrained parameter space. By circumventing complex prompt space searches, our method adaptively bridges domain gaps through feature affinity modeling. For example, our approach achieves gains over TPT~\cite{shu2022test} of 10.99\% and 7.71\% across two CLIP visual backbones, respectively, validating its robustness.

Similarly, cache model-based methods typically adapt solely to the image distribution of the new domain. However, our method unifies query processing for both textual and cached image features, enabling it to adapt to cross-modal information distributions simultaneously. Our approach achieves improvements of 7.62\% and 5.28\% over TDA~\cite{karmanov2024efficient} across two backbones, further supporting the efficacy of our unified strategy.

\subsubsection{Computation Efficiency}

As shown in Tab.~\ref{eff}, we evaluated the efficiency of the Flowers dataset~\cite{nilsback2008automated} on a single Tesla V100S-PCIE-32GB GPU. Note that the number of augmentation views is based on the specifications stated in the original papers for each method. By using a minimal number of image augmentations, we effectively avoided excessive overhead while achieving a 15.47\% improvement over the original CLIP. Additionally, our supplementary material includes analysis and experiments on accelerating inference, highlighting further optimization potential.

\subsection{Ablation Study}

\begin{table}[t]
  \centering
  \caption{Graph ablation with CLIP ViT-B/16 and DINOv2 ViT-S/14. CSS denotes the prediction from the CLIP Shared Semantic Graph, while AFV represents the prediction from the Auxiliary Fine-grained Visual Graph.}
  \resizebox{\columnwidth}{!}{%

    \begin{tabular}{ccccccc}
    \toprule
    \multirow{2}[4]{*}{CLIP} & \multirow{2}[4]{*}{CSS} & \multirow{2}[4]{*}{AFV} & \multicolumn{2}{c}{ImageNet-Val~\cite{deng2009imagenet}} & \multicolumn{2}{c}{Flower102~\cite{nilsback2008automated}} \\
\cmidrule{4-7}          &       &       & \multicolumn{1}{c}{Top-1 Accuracy (\%)} & \multicolumn{1}{c}{Gain (\%)} & Top-1 Accuracy (\%) & Gain (\%) \\
    \midrule
    \checkmark     &       &       & 69.91  & \multicolumn{1}{c}{-} & 72.64  & - \\
          & \checkmark     &       & 71.88  & 1.97  & 74.46  & 1.82  \\
          &       & \checkmark     & 67.05  & -2.86 & 77.71  & 5.07  \\
    \checkmark     & \checkmark     &       & 72.06  & 2.15  & 75.07  & 2.43  \\
    \checkmark     &       & \checkmark     & 73.16  & 3.25  & 79.74  & 7.10  \\
    \checkmark     & \checkmark     & \checkmark     & \textbf{73.79}  & \textbf{3.88}  & \textbf{81.61}  & \textbf{8.97}  \\
    \bottomrule
    \end{tabular}%

      } 
  \label{Components}%
    \vspace{-0.6cm}
  
\end{table}%

\subsubsection{Ablation of Graph Type}

We evaluated the performance of different graphs on ImageNet Val~\cite{deng2009imagenet} and Flowers102~\cite{nilsback2008automated}. As shown in the Tab.~\ref{Components}, when CSS and AFV prediction are fused with the original CLIP predictions, significant improvements are observed, with gains of 2.15\%, 3.25\%, and 2.43\%, 7.10\%, respectively. This indicates that each graph contributes to enhancing the generalization ability of CLIP. Moreover, the performance is optimized when CSS and AFV are used in conjunction, achieving improvements of 3.88\% and 8.97\%, which validates that the different target feature spaces enhance the perception of shared modal features and fine-grained visual capabilities. However, we note a -2.86\% degradation when using AFV alone on ImageNet-Val, as relying solely on cached DINOv2 visual features is insufficient to accurately describe new images with domain gaps.

\subsubsection{Ablation of Auxiliary Visual Encoder}

Introducing an auxiliary visual encoder with various sizes consistently improves model accuracy, as shown in Tab.~\ref{tab:encoder}. Specifically, each encoder boosts accuracy over the baseline (69.91\%), with gains of 3.67\% (ViT-S/14), 6.82\% (ViT-B/14), and 7.63\% (ViT-L/14).  Furthermore, the Tab.~\ref{tab:encoder} highlights a clear trade-off between performance gains and additional parameters. The smallest encoder, ViT-S/14, with only 21M extra parameters, shows modest improvement, while the largest, ViT-L/14, requires 300M additional parameters for the highest gain. 

\subsubsection{Ablation of Logits Weights}

\label{betas}

\begin{table}[t]
  \centering
  \caption{Ablation of various auxiliary visual encoder in ImageNet Val~\cite{deng2009imagenet} with CLIP ViT-B/16.}
    \resizebox{\columnwidth}{!}{%
    \begin{tabular}{cccc}
    \toprule
    DINOv2 Backbone & Extra Parameters & Top-1 Accuracy (\%)   & Gain (\%) \\
    \midrule
    -     & -     & 69.91  & - \\
    ViT-S/14 & 21 M  & 73.79  & 3.88  \\
    ViT-B/14 & 86 M  & 76.73  & 6.82  \\
    \rowcolor[rgb]{ .949,  .949,  .949} ViT-L/14 & 300 M & \textbf{77.54}  & \textbf{7.63}  \\
    \bottomrule
    \end{tabular}%
    }
  \label{tab:encoder}%
\end{table}%

\begin{figure}[t] 
  \centering
  \begin{subfigure}[t]{0.225\textwidth} 
    \centering
    \includegraphics[width=\linewidth]{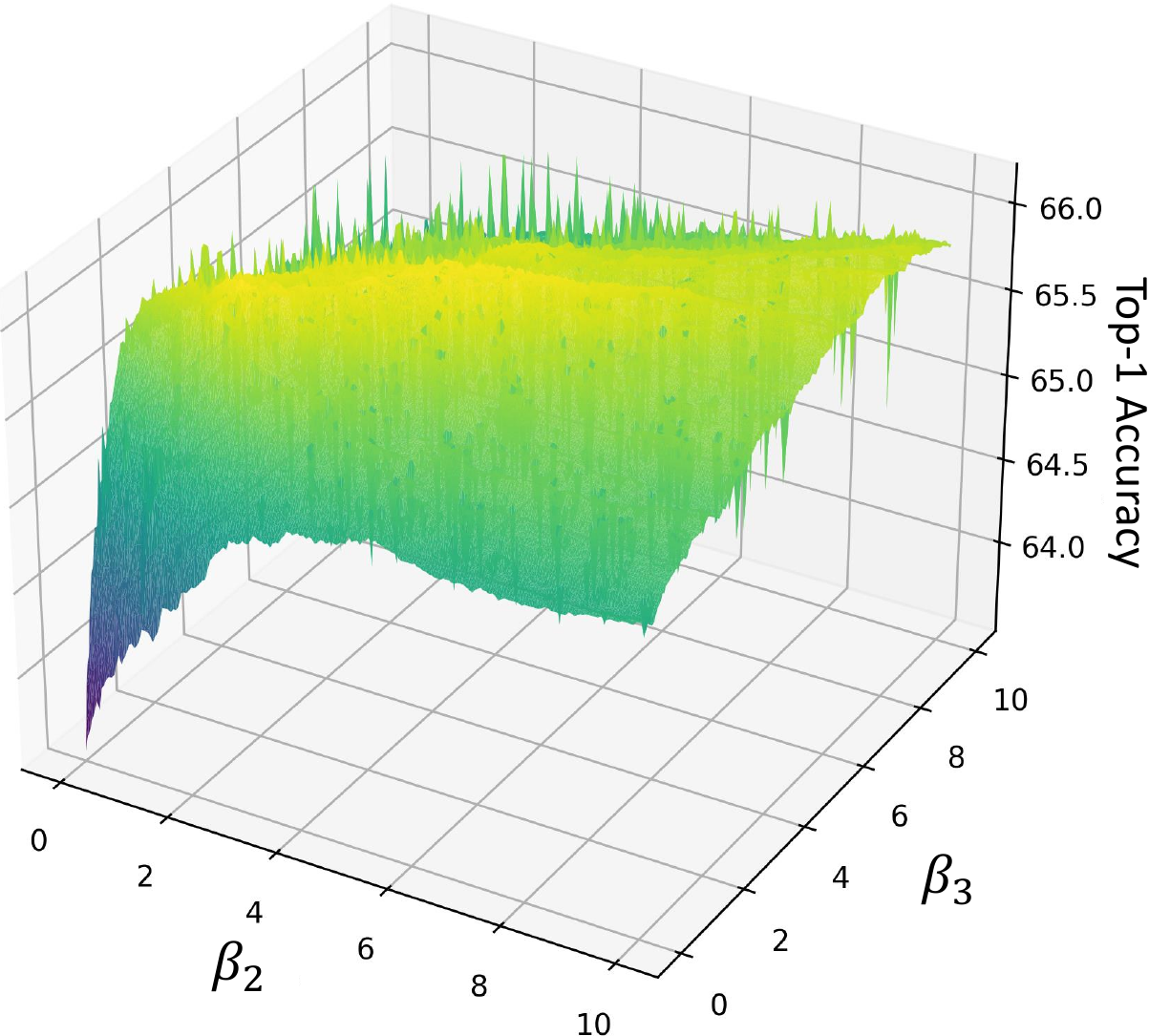} 
    \caption{ImageNet-V2~\cite{recht2019imagenet}}
    \label{combined_images1}
  \end{subfigure}%
  \hspace{0.02\textwidth} 
  \begin{subfigure}[t]{0.225\textwidth} 
    \centering
    \includegraphics[width=\linewidth]{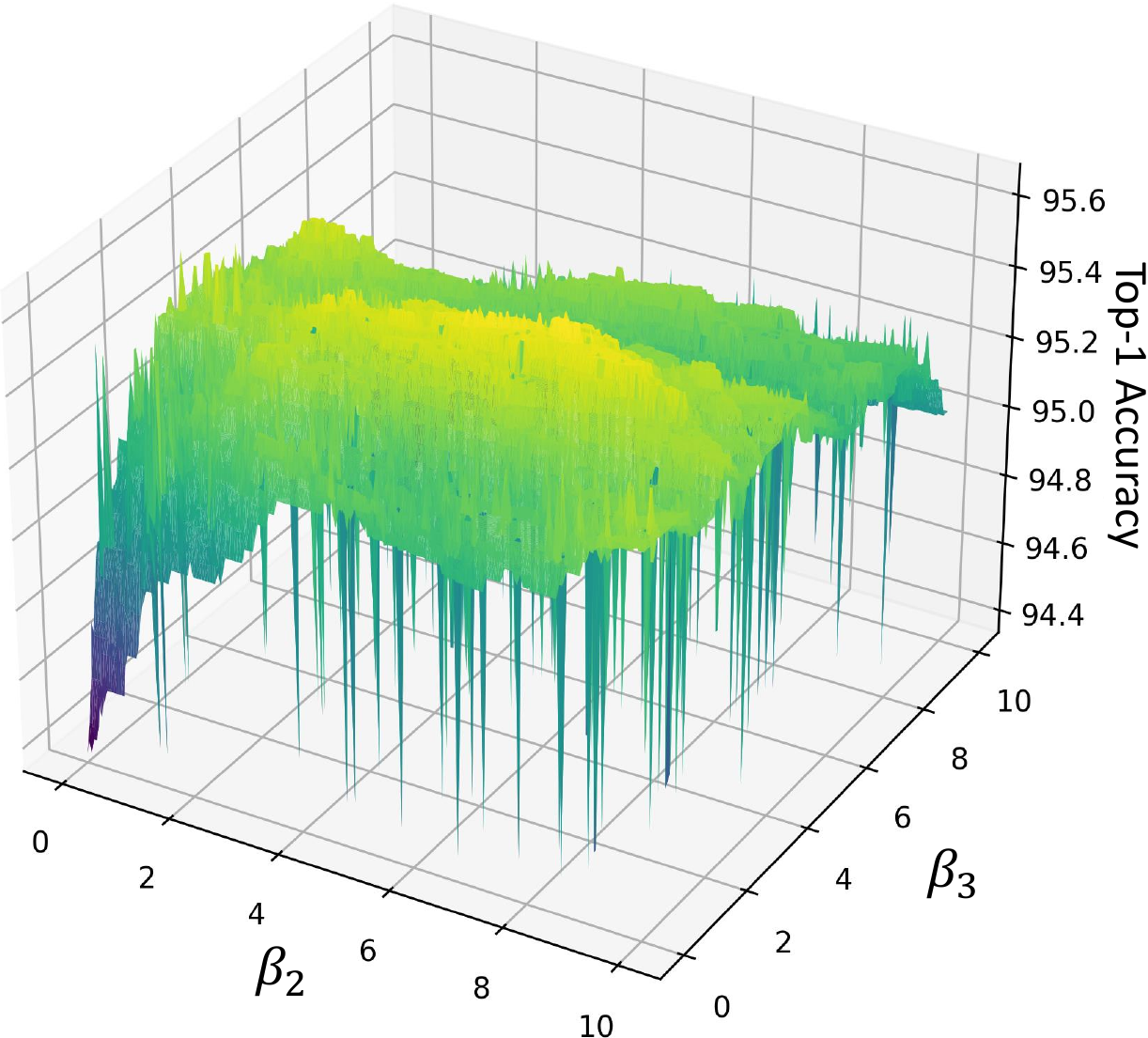} 
    \caption{Caltech101~\cite{fei2004learning}}
    \label{combined_images2}
  \end{subfigure}
  \caption{Ablation of $\beta_2$, $\beta_3$ with CLIP ViT-B/16 and DINOv2 ViT-S/14.}
  \label{combined_images}
  \vspace{-0.5cm}
\end{figure}
To verify the robustness of different predictive logits on the final prediction in Eq.~\ref{add eq}, we fixed the value of $\beta_1$ at 1 and set the step size to 0.05, varying $\beta_2$ and $\beta_3$ from 0 to 10 to observe their impact on model prediction accuracy. As shown in Fig.~\ref{combined_images}, we visualized this on two representative subsets ImageNet-V2~\cite{recht2019imagenet} and Caltech101~\cite{fei2004learning}. As shown, increasing either $\beta_2$ or $\beta_3$ significantly enhances accuracy, indicating that both semantic granularities independently improve generalization. These parameters exhibit strong synergistic properties, mutually promoting the fusion of complementary semantics and thereby achieving superior performance. The model demonstrates robustness, maintaining high accuracy even with perturbations around the optimal $\beta$ values, suggesting that performance is resilient to variations in these hyperparameters.

\subsubsection{Ablation of Components}

\begin{figure}[t] %
  \centering
  \begin{subfigure}[t]{0.225\textwidth} %
    \centering
    \includegraphics[width=\linewidth]{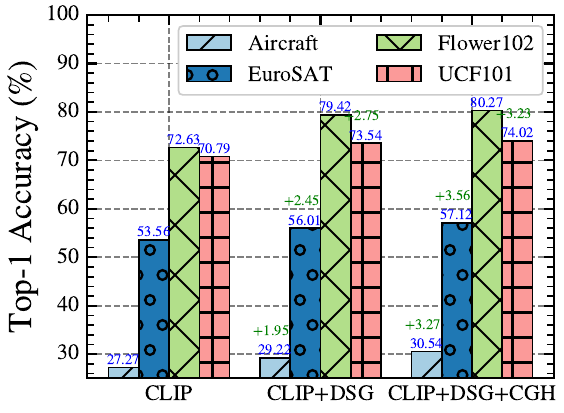} %
    \caption{Ablation of key modules.}
    \label{components1}
  \end{subfigure}%
  \hspace{0.02\textwidth} 
  \begin{subfigure}[t]{0.225\textwidth} %
    \centering
    \includegraphics[width=\linewidth]{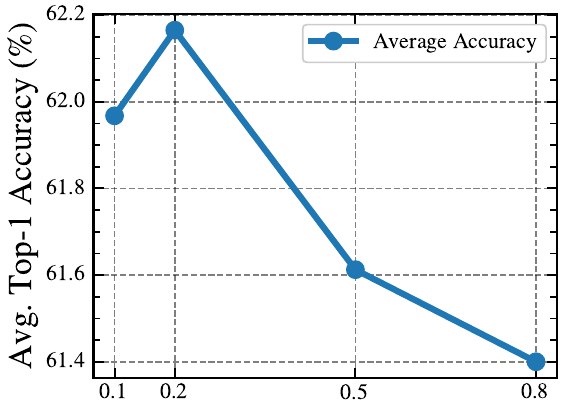} %
    \caption{Ablation of $r$.}
    \label{components2}
  \end{subfigure}
  \caption{Ablation of components with CLIP-ViT-B16.}
  \label{components}
\end{figure}

In Fig.~\ref{components} (a), CLIP+DSG is equivalent to directly computing the similarity between test samples and class nodes within the DSG. Although DSG independently enhances model generalization, CGH highlights the importance of selecting an appropriate $r$ to enhance the model's robustness by effectively filtering out noise and outliers, thereby improving prediction accuracy. As shown in Fig.~\ref{components} (b), we varied the selection ratio $r$ of nearest anchors across multiple datasets to evaluate the impact of different components on model performance. The results indicate that the model achieves optimal performance at $r=0.2$ with both DSG and CGH. This suggests that the similarity between test samples and diverse hyper-classes is more robust than that between samples and normal classes. 

\subsubsection{Visualization}
As shown in Fig.~\ref{fig:attn}, DINOv2’s vision-contrastive encoder outperforms CLIP in capturing key objects. Leveraging finer-grained features, our method excels in recognizing \enquote{hard} classes, accurately distinguishing similar pets (e.g., Pomeranian vs. Japanese Chin) with higher confidence.

\vspace{-0.2cm}
\section{Conclusion}
\label{sec:Conclusion}

In CLIP test-time adaptation, we focus on exploring the potential of the cache-based method. As for how to refine cache, we introduced a \textit{Dual Semantics Graph} to explore inter- and cross-modal affinities with various semantics. As for how to query cache, we introduce \textit{Clique Guided Hyper-classes} within dual graphs to enhance the selection of correlated classes for robust predictions. Our method outperforms SOTA on multiple benchmarks, demonstrating strong zero-shot capabilities, while revealing the significant potential of structured semantic integration for robust visual understanding. \textbf{Limitation:} The current clique search methodology incurs non-trivial time complexity, imposing an additional computational burden on COSMIC. To address this, we aim to explore dual graph sparsification techniques for accelerated search in future iterations.

\section*{Acknowledgment}  

This work was supported by the following funding sources: the National Key Research and Development Project of China (Grant No. 2023YFF0905502), the National Natural Science Foundation of China (Grant Nos. 92467204 and 62472249), the Shenzhen Science and Technology Program (Grant Nos. JCYJ20220818101014030 and KJZD20240903102300001), the Scientific Research Startup Fund of TSIGS (Project No. QD2022004C), the National Natural Science Foundation of China (Grant No. W2432041), and the Natural Science Foundation of Top Talent of SZTU (Grant No. GDRC202413).
{
    \small
\bibliographystyle{ieeenat_fullname}
    \bibliography{main}
}



\clearpage
\setcounter{page}{1}
\maketitlesupplementary

\section{Overview}

This supplementary material provides additional experiments, visualizations, and implementation details to support our main paper. The content is organized as follows:

\begin{itemize}

    \item \textbf{Extra Ablation Experiments} (Sec.~\ref{sec:extra_ablation}). We analyze the impact of DINOv2 cache capacity, image augmentation views, and AFV class center calculation methods on our model's performance.
    \item \textbf{Extra Overhead Discussion} (Sec.~\ref{sec:extra_overhead}). We analyze the storage and time complexity of our method, showing its efficiency through reduced dual graph update frequency and detailed complexity approximations.
    \item \textbf{Extra Visualization} (Sec.~\ref{sec:extra_viz}). We present t-SNE visualizations of class and hyper-class distributions, cached features from CLIP and DINOv2, and examples of queried classes to illustrate our method's effectiveness.
    \item \textbf{Extra Implementation Details} (Sec.~\ref{sec:extra_impl}). We provide comprehensive dataset statistics and textual prompts used for various recognition tasks.
\end{itemize}

These materials offer a deeper understanding of our method's overhead, robustness, visual performance, and experimental setup.

\section{Extra Ablation Experiments}
\label{sec:extra_ablation}

\subsection{Ablation of the Capacity of DINOv2 Cache}

We present the performance of COSMIC with different capacities (number of examples stored) in DINOv2's cache in Tab.~\ref{tab:shot}. It shows that increasing the number of stored examples leads to better prediction, but COSMIC achieves reasonable performance even with a smaller cache capacity.

\subsection{Ablation of the Augment Views of Images}

We evaluated the performance of COSMIC with different numbers of augmented views of images, as shown in Tab.~\ref{tab:views}. The results indicate that increasing the number of views enhances prediction. Specifically, COSMIC achieves its highest Top-1 accuracy gain (2.02\%) with 16 views. However, even with fewer augmented views, COSMIC still performs well and has the advantage of faster inference times. 

\begin{table}[htbp]
  \centering
  \caption{Performance comparison using different cache capacities (number of examples stored) in DINOv2's cache on ImageNet-Val~\cite{deng2009imagenet}. For each test, we use CLIP-RN-50 and DINOv2 ViT-S/14 as our visual encoders.}
  \resizebox{\columnwidth}{!}{%
    \begin{tabular}{lccc}
    \toprule
    Method & \# of Examples Stored & Top-1 Accuracy (\%) & Gain (\%) \\
    \midrule
    CLIP-RN-50 & -     & 66.99  & - \\
    \multirow{5}[1]{*}{Ours} & 1     & 67.10  & 0.11  \\
          & 3     & 68.59  & 1.60  \\
          & 6     & 68.90  & 1.91  \\
          & 8     & 68.92  & 1.93  \\
          & 10    & 68.86  & 1.87  \\
    \bottomrule
    \end{tabular}%
  }
  \label{tab:shot}%
\end{table}%

\begin{table}[htbp]
  \centering
  \caption{Performance comparison using different augment views of images on ImageNet-Val~\cite{deng2009imagenet}. We use CLIP-RN-50 and DINOv2 ViT-S/14 for each test as our visual encoders.}
      \resizebox{\columnwidth}{!}{%
    \begin{tabular}{lccc}
    \toprule
    Method & \# of Augment Views & Top-1 Accuracy (\%) & Gain (\%) \\
    \midrule
    CLIP-RN-50 & -     & 66.99  & - \\
    \multirow{6}[1]{*}{Ours} & 1     & 68.37  & 1.47  \\
          & 2     & 68.59  & 1.69  \\
          & 4     & 68.79  & 1.89  \\
          & 8     & 68.87  & 1.97  \\
          & 16    & 68.92  & 2.02  \\
          & 32    & 68.90  & 2.00  \\
    \bottomrule
    \end{tabular}%
    }
  \label{tab:views}%
\end{table}%

\subsection{Ablation of Calculation of AFV Class Center}

To investigate the impact of various class center calculation methods in the Auxiliary Fine-grained Visual space on performance, we conducted a comparative analysis. Tab.~\ref{tab:afv_center} shows our method significantly improves upon the CLIP-RN-50 baseline using both average and attention-weighted AFV class centers. The average method achieves the highest Top-1 accuracy gain (1.91\%), slightly surpassing the attention-weighted method (1.62\%) and the EMA method (1.60\%). This suggests that equal consideration of all cached features may better capture class-level representations. The slight performance decrease (-0.03\%) of the EMA method without entropy-based selection emphasizes the importance of careful feature selection. These results highlight the critical role of AFV class center calculation in leveraging cached features, with the simple averaging method emerging as the preferred choice due to its effectiveness and simplicity.

\begin{table}[htbp]
  \centering
  \caption{Performance comparison using different AFV class center calculations on ImageNet-Val~\cite{deng2009imagenet}. CLIP-RN-50 and DINOv2 ViT-S/14 are used as visual encoders. ``Average" means the centroid of cached features for each class. ``Attn weighted" means the weighted average of cached features for each class, with weights being the attention scores between the test feature and cached features. ``EMA" means the exponential moving average of historical test features. ``EN" means the prediction entropy-based selection of features.}
  \small
          \resizebox{\columnwidth}{!}{%
  \begin{tabular}{llcc}

    \toprule
    Method & AFV Center  & Top-1 Accuracy (\%) & Gain (\%) \\
    \midrule
    CLIP-RN-50 & -     & 66.99 & - \\
    \multirow{3}{*}{Ours} & Average & 68.90 & 1.91 \\
          & Attn weighted & 68.61 & 1.62 \\
          & EMA   & 68.59  & 1.60  \\
          & EMA w/o EN & 66.96 & -0.03 \\
    \bottomrule
      
  \end{tabular}%
  }

  \label{tab:afv_center}%
\end{table}%

\begin{figure*}[t]
  \centering
  \includegraphics[width=\textwidth]{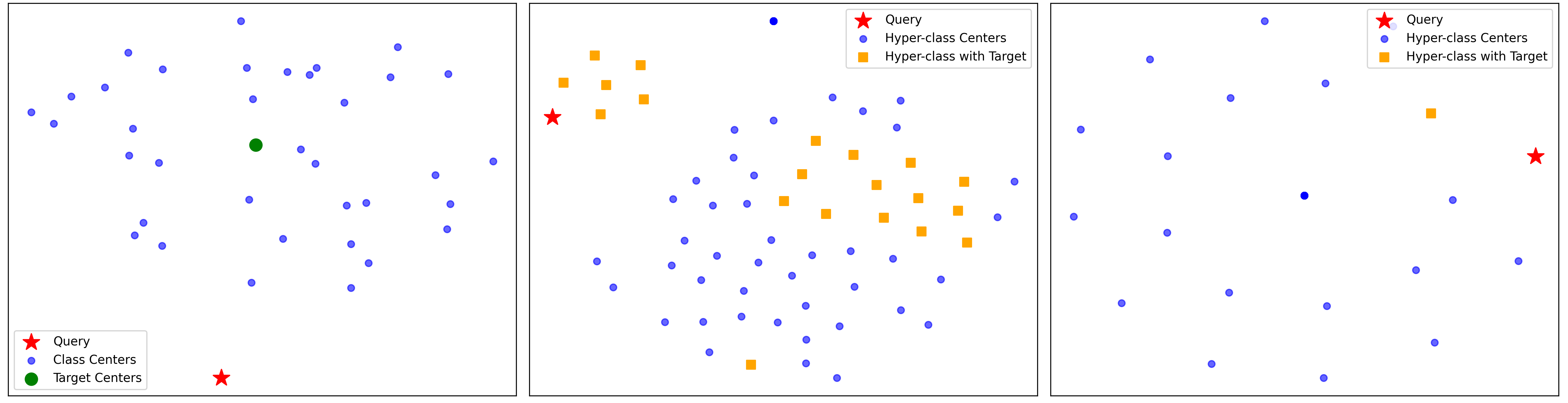}
  \caption{With a sample from Pets dataset~\cite{parkhi2012cats}, we implement t-SNE visualization of test features querying Textual Class Centers (\textbf{left}), CLIP Shared Semantics Hyper-class Centers (\textbf{middle}), and Auxiliary Fine-grained Visual Hyper-class Centers (\textbf{right}). ``Target" denotes the ground-truth label. CLIP-ViT-B/16 and DINOv2 ViT-L/14 serve as visual encoders.}
  \label{fig:sup_class}
\end{figure*}

\section{Extra Overhead Discussion}
\label{sec:extra_overhead}
As shown in Tab.~\ref{tab2}, time cost can be reduced in real applications by decreasing the frequency of dual graph updates—such as every 50 steps—while still achieving SOTA. \textbf{[Storage]}: Constructing additional graph structures only requires $\mathcal{O}(K^2)$ space to store the adjacency matrix. Additionally, storing extra visual features only requires (class\_num ($K$) + clique\_num) $\times$ cache\_size ($n$) $\times$ feat\_dim ($d_i$) space, which is highly efficient with pytorch tensor. The approximated {total storage/sample}: $\mathcal{O}((d_1+d_2)n K +K^2+\mathrm{clique\_num} \times(d_1+d_2))$. \textbf{[Time]}: Time complexity $b(K - b)3^{(b/3)}$ of maximal clique search is presented in main text. The approximated {total time/sample}: $\max\left(\mathcal{O}(\mathrm{CLIP}),\mathcal{O}(\mathrm{DINOv2})\right)+\mathcal{O}(d_1 (2K)^2+d_2 K^2+b(K - b)3^{(b/3)}+nK\times \mathrm{clique\_num})$ where $b$ is graph degeneracy.

\begin{table*}[htbp]
  \footnotesize 
  \centering
  \caption{We use CLIP ViT-B-16 and DINOv2 ViT-S/14 as the backbone, updating the dual graph every 50 steps to show the average time \& storage overhead per test sample.}
  \resizebox{\textwidth}{!}{
    \begin{tabular}{llccccc}
      \toprule
      \multirow{2}{*}{Test} & \multirow{2}{*}{Type} & \multicolumn{1}{c}{\multirow{2}{*}{CLIP Inference}} & \multicolumn{1}{c}{\multirow{2}{*}{TDA Overhead}} & \multicolumn{3}{c}{COSMIC Overhead} \\
            &       &       &       & \multicolumn{1}{c}{DINOv2 Inference} & \multicolumn{1}{c}{CLIP Graph} & \multicolumn{1}{c}{DINOv2 Graph} \\
      \midrule
      \multirow{3}{*}{Flower102~\cite{nilsback2008automated}} & Time {(ms)} & 12.52  & 8.42  & 10.65  & 5.37  & 3.75  \\
    & Storage {(mb)}  & 147.87  & 40.93  & 42.84  & 0.43  & 0.15  \\
    & Top-1 Acc {(\%)}  & 72.76  & 75.11  & -  & 77.10  & 80.92  \\
    \midrule
      \multirow{3}{*}{Ucf101~\cite{khurram2012ucf101}} & Time {(ms)} & 17.94  & 10.99  & 12.20  & 6.31  & 4.32  \\
    & Storage {(mb)} & 147.91  & 40.61  & 74.09  & 1.62  & 1.46  \\
    & Top-1 Acc {(\%)} & 94.36  & 94.40  & -  & 94.77  & 95.33  \\
      \bottomrule
    \end{tabular}
    }
  \label{tab2}
  
\end{table*}

\section{Extra Visualization}
\label{sec:extra_viz}
\subsection{Distribution of Classes and Hyper-classes}

To showcase the effectiveness of our method during the cold-start phase, we visualize the distribution of randomly selected test samples from the first 100 tests in the Pets dataset~\cite{parkhi2012cats} across three query spaces: Textual Class Centers, CLIP Shared Semantics Hyper-class Centers, and Auxiliary Fine-grained Visual Hyper-class Centers. We employ t-SNE to reduce the dimensionality of the high-dimensional features. As illustrated in Fig.~\ref{fig:sup_class}, hyper-classes exhibit a more uniform distribution in the feature space. Notably, when ground truth (GT) feature centers are obscured by neighboring points, the Hyper-class Centers containing the GT target are more readily queried by test samples, resulting in improved prediction accuracy.

\subsection{T-SNE of Cached Features from CLIP \& DINOv2}

\begin{figure}[htbp]
    \centering
    \begin{subfigure}[b]{\columnwidth}
        \centering
        \includegraphics[width=\textwidth]{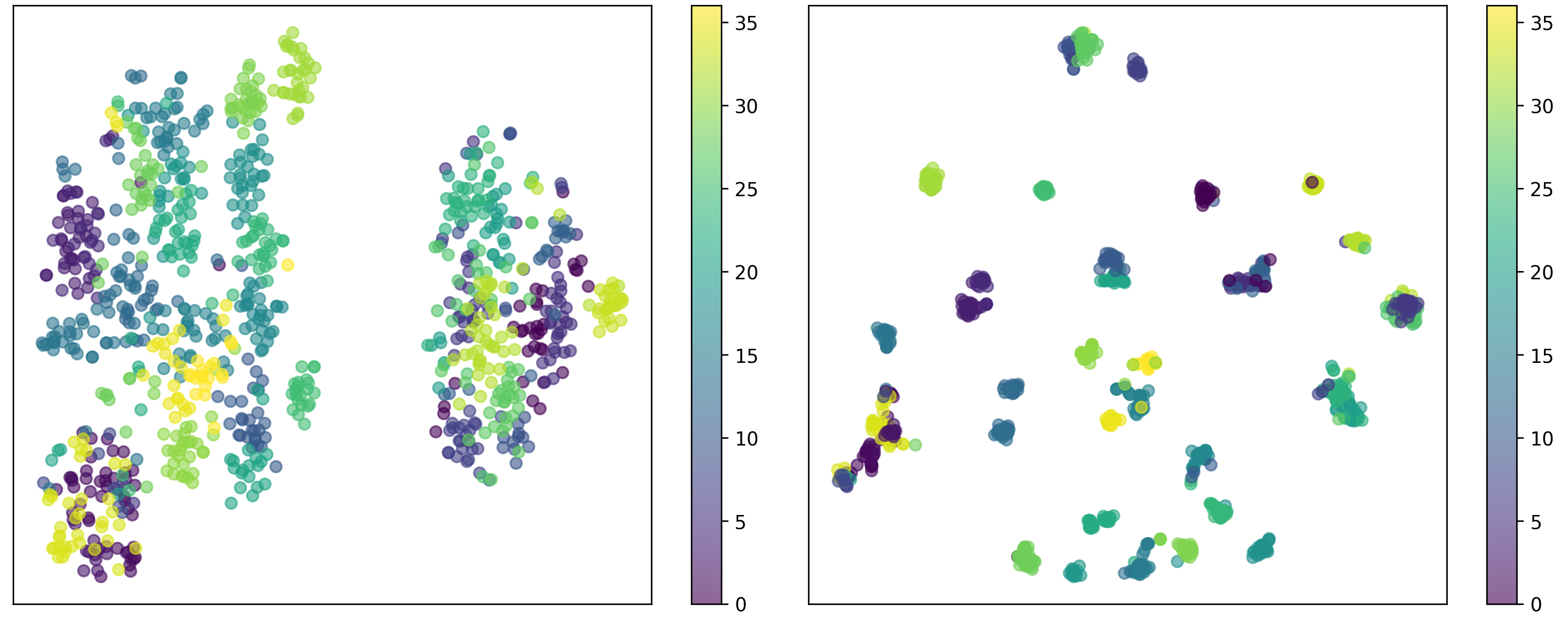}
        \caption{Pets~\cite{parkhi2012cats}}
        \label{fig:image1}
    \end{subfigure}
    
    \vspace{1em}  
    
    \begin{subfigure}[b]{\columnwidth}
        \centering
        \includegraphics[width=\textwidth]{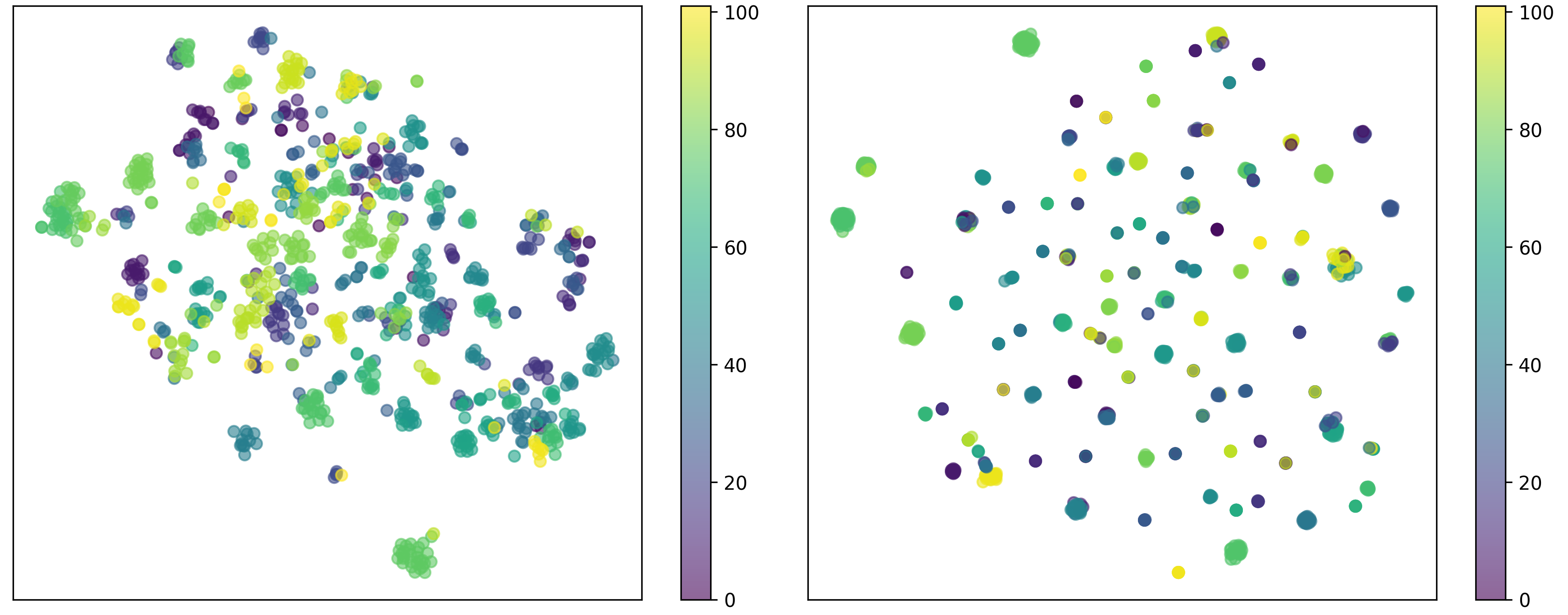}
        \caption{Flower102~\cite{nilsback2008automated}}
        \label{fig:image2}
    \end{subfigure}
    \caption{Distribution of cached visual features from CLIP (\textbf{left}) and DINOv2 (\textbf{right}) caches. The capacity of both caches are set to 50 and we capture the distribution in 1000 test iterations. CLIP-ViT-B/16 and DINOv2 ViT-L/14 serve as visual encoders.}
    \label{fig:ts_ca}
\end{figure}

In Fig.~\ref{fig:ts_ca}, we visualize the cached visual features from CLIP and DINOv2 caches after testing on various subsets of data using t-SNE. We observe that features of the same class (same color) in the DINOv2 cache are more clustered, especially during the cold-start phase, where it exhibits more distinctive class clustering and effectively mitigates overlap between similar categories, thereby facilitating fine-grained visual feature retrieval.

\subsection{Samples of Queried Classes}

\begin{figure*}[t] 
  \centering
  \includegraphics[width=\textwidth]{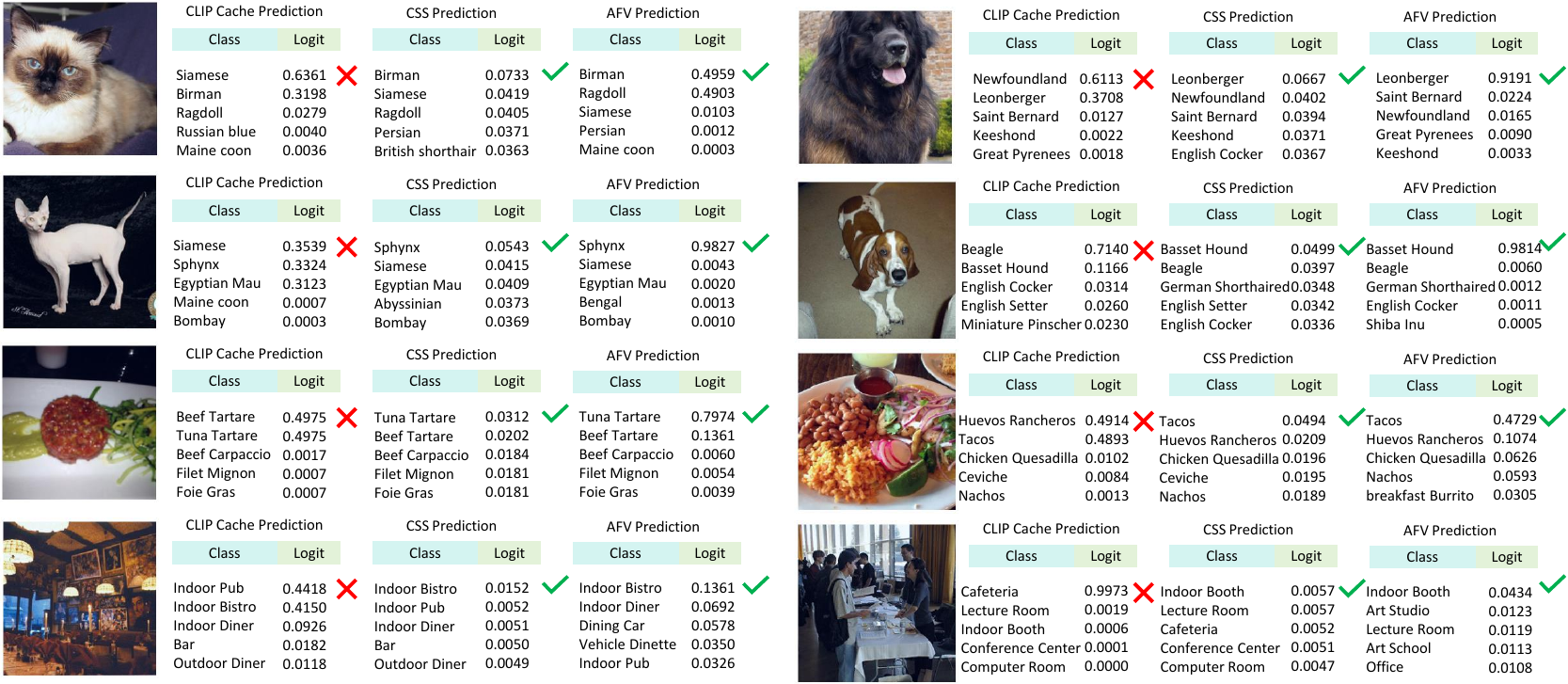} 
  \caption{Samples of queried classes with clip feature cache, CSS Graph, and AFV Graph respectively. For each test, CLIP-ViT-B/16 and DINOv2 ViT-L/14 are used as visual encoders.}
    \label{fig:sup_samples}
\end{figure*}

Fig.~\ref{fig:sup_samples} illustrates the enhanced performance achieved by querying hyper-classes within the CLIP Shared Semantics and Auxiliary Fine-grained Visual graphs, as opposed to the conventional approach of querying classes in the naive CLIP cache. Both graphs leverage the structured relationships and hierarchical organization of hyper-classes, enabling more precise and contextually relevant retrieval of semantic information.

\section{Extra Implementation Details}
\label{sec:extra_impl}

\subsection{Dataset Details}

In Tab.~\ref{tab:datasets}, we present detailed statistics for each dataset used in our experiments, including the number of classes, test set sizes, and their respective target tasks.

\begin{table*}[htbp]
  \centering
  \caption{Dataset Summary for Various Recognition Tasks. Note that we evaluate test datasets for all benchmarks.}
  \resizebox{\textwidth}{!}{%
    \begin{tabular}{lccccl}
    \toprule
    Dataset & Classes & Train size & Validation size & Test size & Target Task \\
    \midrule
    Caltech101~\cite{fei2004learning} & 100 & 4,128 & 1,649 & 2,465 & Object recognition \\
    DTD~\cite{cimpoi2014describing} & 47 & 2,820 & 1,128 & 1,692 & Texture recognition \\
    EuroSAT~\cite{helber2019eurosat} & 10 & 13,500 & 5,400 & 8,100 & Satellite image recognition \\
    FGVCAircraft~\cite{maji2013fine} & 100 & 3,334 & 3,333 & 3,333 & Fine-grained aircraft recognition \\
    Flowers102~\cite{nilsback2008automated} & 102 & 4,093 & 1,633 & 2,463 & Fine-grained flowers recognition \\
    Food101~\cite{bossard2014food} & 101 & 50,500 & 20,200 & 30,300 & Fine-grained food recognition \\
    OxfordPets~\cite{parkhi2012cats} & 37 & 2,944 & 736 & 3,669 & Fine-grained pets recognition \\
    StanfordCars~\cite{krause20133d} & 196 & 6,509 & 1,635 & 8,041 & Fine-grained car recognition \\
    SUN397~\cite{xiao2010sun} & 397 & 15,880 & 3,970 & 19,850 & Scene recognition \\
    UCF101~\cite{khurram2012ucf101} & 101 & 7,639 & 1,898 & 3,783 & Action recognition \\
    \midrule
    ImageNet~\cite{deng2009imagenet} & 1,000 & 1.28M & - & 50,000 & Object recognition \\
    ImageNet-A~\cite{hendrycks2021natural} & 200 & - & - & 7,500 & Robustness of adversarial attack \\
    ImageNet-V2~\cite{recht2019imagenet} & 1,000 & - & - & 10,000 & Robustness of collocation \\
    ImageNet-R~\cite{hendrycks2021many} & 200 & - & - & 30,000 & Robustness of multi-domains \\
    ImageNet-Sketch~\cite{wang2019learning} & 1,000 & - & - & 50,889 & Robustness of sketch domain \\
    \bottomrule
    \end{tabular}%
  }
  \label{tab:datasets}
\end{table*}

\subsection{Textual Prompts Details}

Tab.~\ref{tab:prompts} outlines the prompt formats for various visual recognition datasets. These prompts guide the model in identifying specific objects or scenes within each class, with tailored designs for optimal performance. This variation enhances the model's generalization and accuracy.

\begin{table*}[htbp]
  \centering
  \caption{Textual Prompts for Various Recognition Tasks. The left column lists the dataset names, while the right column provides the prompt templates for each dataset, with empty curly braces {} representing the class placeholder.}
  \resizebox{\textwidth}{!}{%
    \begin{tabular}{lc}
    \toprule
    Dataset & Prompts \\
    \midrule
    Caltech101~\cite{fei2004learning} & “a photo of a \{\}.” \\
    DTD~\cite{cimpoi2014describing} & “\{\} texture.” \\
    EuroSAT~\cite{helber2019eurosat} & “a centered satellite photo of \{\}.” \\
    FGVCAircraft~\cite{maji2013fine} & “a photo of a \{\}, a type of aircraft.” \\
    Flowers102~\cite{nilsback2008automated} & “a photo of a \{\}, a type of flower.” \\
    Food101~\cite{bossard2014food} & “a photo of \{\}, a type of food.” \\
    OxfordPets~\cite{parkhi2012cats} & “a photo of a \{\}, a type of pet.” \\
    StanfordCars~\cite{krause20133d} & “a photo of a \{\}, a type of car.” \\
    SUN397~\cite{xiao2010sun} & “a bad photo of the \{\}.”, “a \{\} in a video game.”, “a origami \{\}.”, “a photo of the small \{\}.”, “art of the \{\}.”, “a photo of the large \{\}.”, “itap of a \{\}.” \\
    UCF101~\cite{khurram2012ucf101} & “a photo of a person doing \{\}.” \\
    
    \midrule
    ImageNet~\cite{deng2009imagenet} & “a bad photo of the \{\}.”, “a \{\} in a video game.”, “a origami \{\}.”, “a photo of the small \{\}.”, “art of the \{\}.”, “a photo of the large \{\}.”, “itap of a \{\}.” \\
    ImageNet-A~\cite{hendrycks2021natural} & “a bad photo of the \{\}.”, “a \{\} in a video game.”, “a origami \{\}.”, “a photo of the small \{\}.”, “art of the \{\}.”, “a photo of the large \{\}.”, “itap of a \{\}.” \\
    ImageNet-V2~\cite{recht2019imagenet} & “a bad photo of the \{\}.”, “a \{\} in a video game.”, “a origami \{\}.”, “a photo of the small \{\}.”, “art of the \{\}.”, “a photo of the large \{\}.”, “itap of a \{\}.” \\
    ImageNet-R~\cite{hendrycks2021many} & “a bad photo of the \{\}.”, “a \{\} in a video game.”, “a origami \{\}.”, “a photo of the small \{\}.”, “art of the \{\}.”, “a photo of the large \{\}.”, “itap of a \{\}.” \\
    ImageNet-Sketch~\cite{wang2019learning} & “a bad photo of the \{\}.”, “a \{\} in a video game.”, “a origami \{\}.”, “a photo of the small \{\}.”, “art of the \{\}.”, “a photo of the large \{\}.”, “itap of a \{\}.” \\
    
    \bottomrule
    \end{tabular}%
  }
  \label{tab:prompts}
\end{table*}

\end{document}


\section*{Derivation of TDA's Prediction Mechanism}

\subsection*{1. Original Formula}

The prediction logits for a test sample \( f_{\mathrm{test}} \) in Tip-Adapter (TDA) are defined as:
\begin{equation}
    P_{\mathrm{cache}}(f_{\mathrm{test}}) = A(f_{\mathrm{test}} \mathbf{F}_{\mathrm{train}}^\top) \mathbf{L}_{\mathrm{train}},
\end{equation}
where:
\begin{itemize}
    \item \( f_{\mathrm{test}} \in \mathbb{R}^D \): The test sample's visual feature vector.
    \item \( \mathbf{F}_{\mathrm{train}} \in \mathbb{R}^{N \times D} \): The cached feature matrix, where each row corresponds to the \(N\) training sample features.
    \item \( \mathbf{L}_{\mathrm{train}} \in \mathbb{R}^{N \times C} \): The one-hot label matrix, where each row is the one-hot representation of the corresponding training sample's label among \(C\) classes.
    \item \( A(\cdot) \): A scaling function applied to the similarity scores (e.g., softmax normalization or a temperature scaling).
\end{itemize}

---

\subsection*{2. Simplification by Removing \( A(\cdot) \)}

The scaling function \( A(\cdot) \) only modifies the values of the resulting similarity matrix without affecting its dimensionality or structure. Therefore, for the purpose of derivation, we can temporarily ignore \( A(\cdot) \). This simplifies the formula to:
\begin{equation}
    P_{\mathrm{cache}}(f_{\mathrm{test}}) = (f_{\mathrm{test}} \mathbf{F}_{\mathrm{train}}^\top) \mathbf{L}_{\mathrm{train}}.
\end{equation}

---

\subsection*{3. Associativity of Matrix Multiplication}

The above formula involves three terms combined through matrix multiplication: \( f_{\mathrm{test}} \), \( \mathbf{F}_{\mathrm{train}}^\top \), and \( \mathbf{L}_{\mathrm{train}} \). To simplify further, we need to examine the associativity of matrix multiplication.

\paragraph{Matrix Multiplication Associativity:}
For matrices \( \mathbf{A} \), \( \mathbf{B} \), and \( \mathbf{C} \) with compatible dimensions, the following property holds:
\begin{equation}
    (\mathbf{A} \mathbf{B}) \mathbf{C} = \mathbf{A} (\mathbf{B} \mathbf{C}).
\end{equation}

This property allows us to group the terms in any order, provided the dimensions of the matrices align. Let us verify the dimensions of the terms in our formula:
\begin{itemize}
    \item \( f_{\mathrm{test}} \in \mathbb{R}^D \).
    \item \( \mathbf{F}_{\mathrm{train}}^\top \in \mathbb{R}^{D \times N} \).
    \item \( \mathbf{L}_{\mathrm{train}} \in \mathbb{R}^{N \times C} \).
\end{itemize}

The intermediate terms are computed as follows:
\begin{enumerate}
    \item \( f_{\mathrm{test}} \mathbf{F}_{\mathrm{train}}^\top \in \mathbb{R}^N \): This computes the similarity between the test sample and all cached training features.
    \item \( \mathbf{F}_{\mathrm{train}}^\top \mathbf{L}_{\mathrm{train}} \in \mathbb{R}^{D \times C} \): This computes the aggregated class-level features (explained below).
\end{enumerate}

Using the associativity property, we can group the terms as:
\begin{equation}
    P_{\mathrm{cache}}(f_{\mathrm{test}}) = f_{\mathrm{test}} (\mathbf{F}_{\mathrm{train}}^\top \mathbf{L}_{\mathrm{train}}).
\end{equation}
This regrouping is valid because the dimensions align correctly:
\begin{itemize}
    \item \( \mathbf{F}_{\mathrm{train}}^\top \mathbf{L}_{\mathrm{train}} \in \mathbb{R}^{D \times C} \),
    \item \( f_{\mathrm{test}} (\mathbf{F}_{\mathrm{train}}^\top \mathbf{L}_{\mathrm{train}}) \in \mathbb{R}^C \),
\end{itemize}
which matches the required output dimensionality: a vector of logits for \( C \) classes.

---

\subsection*{4. The Role of \( \mathbf{F}_{\mathrm{train}}^\top \mathbf{L}_{\mathrm{train}} \)}

The intermediate matrix \( \mathbf{F}_{\mathrm{train}}^\top \mathbf{L}_{\mathrm{train}} \) is central to understanding the mechanism. Let us analyze it in detail.

1. **Structure of \( \mathbf{L}_{\mathrm{train}} \):**
   Each column \( \mathbf{L}_{\mathrm{train}}[:, c] \in \mathbb{R}^N \) of \( \mathbf{L}_{\mathrm{train}} \) is a binary indicator vector for class \( c \):
   \[
   \mathbf{L}_{\mathrm{train}}[i, c] =
   \begin{cases}
       1 & \text{if training sample } i \text{ belongs to class } c, \\
       0 & \text{otherwise.}
   \end{cases}
   \]

2. **Matrix Multiplication \( \mathbf{F}_{\mathrm{train}}^\top \mathbf{L}_{\mathrm{train}} \):**
   The product \( \mathbf{F}_{\mathrm{train}}^\top \mathbf{L}_{\mathrm{train}} \in \mathbb{R}^{D \times C} \) aggregates the training sample features for each class. Specifically, the \( c \)-th column of \( \mathbf{F}_{\mathrm{train}}^\top \mathbf{L}_{\mathrm{train}} \) is:
   \[
   (\mathbf{F}_{\mathrm{train}}^\top \mathbf{L}_{\mathrm{train}})[:, c] = \sum_{i \in I_c} f_i,
   \]
   where \( I_c = \{i \mid \mathbf{L}_{\mathrm{train}}[i, c] = 1\} \) is the set of indices of training samples belonging to class \( c \), and \( f_i \) is the visual feature of the \( i \)-th training sample.

   In other words, \( \mathbf{F}_{\mathrm{train}}^\top \mathbf{L}_{\mathrm{train}} \) computes the **sum of features for each class**.

3. **Class Centroids:**
   To compute the class centroid \( \mu_c \) for each class \( c \), we divide the sum of features by the number of samples in the class, \( |I_c| \):
   \[
   \mu_c = \frac{1}{|I_c|} \sum_{i \in I_c} f_i.
   \]

   In matrix form, the centroids matrix \( \mathbf{C} \in \mathbb{R}^{D \times C} \) can be expressed as:
   \[
   \mathbf{C} = \mathbf{F}_{\mathrm{train}}^\top \mathbf{L}_{\mathrm{train}} \mathbf{D}^{-1},
   \]
   where \( \mathbf{D} \in \mathbb{R}^{C \times C} \) is a diagonal matrix with \( \mathbf{D}[c, c] = |I_c| \), the number of samples in class \( c \).

---

\subsection*{5. Final Reformulation}

Substituting \( \mathbf{F}_{\mathrm{train}}^\top \mathbf{L}_{\mathrm{train}} \) into the simplified formula, we get:
\begin{equation}
    P_{\mathrm{cache}}(f_{\mathrm{test}}) = f_{\mathrm{test}} \mathbf{C},
\end{equation}
where \( \mathbf{C} \in \mathbb{R}^{D \times C} \) is the matrix of class centroids.

Each element of \( P_{\mathrm{cache}}(f_{\mathrm{test}}) \) corresponds to the similarity between the test sample \( f_{\mathrm{test}} \) and the centroid of class \( c \). These logits are then passed through the softmax function to obtain class probabilities:
\begin{equation}
    P(c \mid f_{\mathrm{test}}) = \frac{\exp(\ell[c])}{\sum_{c'=1}^C \exp(\ell[c'])},
\end{equation}
where \( \ell[c] = f_{\mathrm{test}} \cdot \mu_c \) is the cosine similarity between \( f_{\mathrm{test}} \) and the centroid \( \mu_c \).

---

\subsection*{6. Summary}

The derivation can be summarized as follows:
\begin{itemize}
    \item **Associativity:** The formula can be rearranged using the associativity of matrix multiplication.
    \item **Centroid Calculation:** The term \( \mathbf{F}_{\mathrm{train}}^\top \mathbf{L}_{\mathrm{train}} \) computes the sum of features for each class, and dividing by the class size gives the centroids.
    \item **Logits:** The prediction is the similarity between the test feature and the class centroids.
\end{itemize}